\documentclass[runningheads]{llncs}

\usepackage{eccv}

\usepackage{eccvabbrv}

\usepackage{graphicx}
\usepackage{booktabs}

\usepackage[accsupp]{axessibility}  

\usepackage{hyperref}

\usepackage{orcidlink}

\usepackage{xcolor}
\usepackage{colortbl} 
\usepackage{multicol}
\usepackage{multirow}
\usepackage{makecell}
\usepackage{rotating}
\usepackage{tabularx}
\usepackage{wrapfig}
\usepackage{amssymb}
\usepackage{pifont}
\usepackage{arydshln}

\definecolor{colorFirst}{RGB}{165, 220, 165} 
\definecolor{colorSecond}{RGB}{192, 232, 212}
\definecolor{colorThird}{RGB}{225, 245, 232} 

\newcommand{\cellrankfirst}[1]{\cellcolor{colorFirst}\textbf{#1}}
\newcommand{\cellranksecond}[1]{\cellcolor{colorSecond}#1}
\newcommand{\cellrankthird}[1]{\cellcolor{colorThird}#1}

\begin{document}

\title{Efficient and High-Quality Depth Estimation via Pixel-Space Diffusion with Linear Attention} 

\titlerunning{Depth Estimation via Pixel-Space Diffusion with Linear Attention}

\author{
    Bingde Liu\inst{1, 2} \and
    Wu Ran\inst{1} \and
    Jinglei Zhang\inst{1} \and
    Huanhuan Yuan\inst{1} \and
    Chao Ma\inst{1,}\thanks{Corresponding author.}
}

\authorrunning{B.~Liu et al.}

\institute{
    MoE Key Lab of Artificial Intelligence, Institute of AI, Shanghai Jiao Tong \\ University, Shanghai, China \and
    Zhiyuan College, Shanghai Jiao Tong University, Shanghai, China \\
    \email{\{cylarus1024, bonjourlemonde, zhangjinglei168, yuanhuanhuan, chaoma\}@sjtu.edu.cn} \\
    Code: \url{https://github.com/VISION-SJTU/Lapis} \\
}

\maketitle

\begin{figure}[th]
  \centering
  \includegraphics[
    width=\textwidth,
    trim=3.5cm 2.2cm 2.8cm 2.0cm,
    clip
  ]{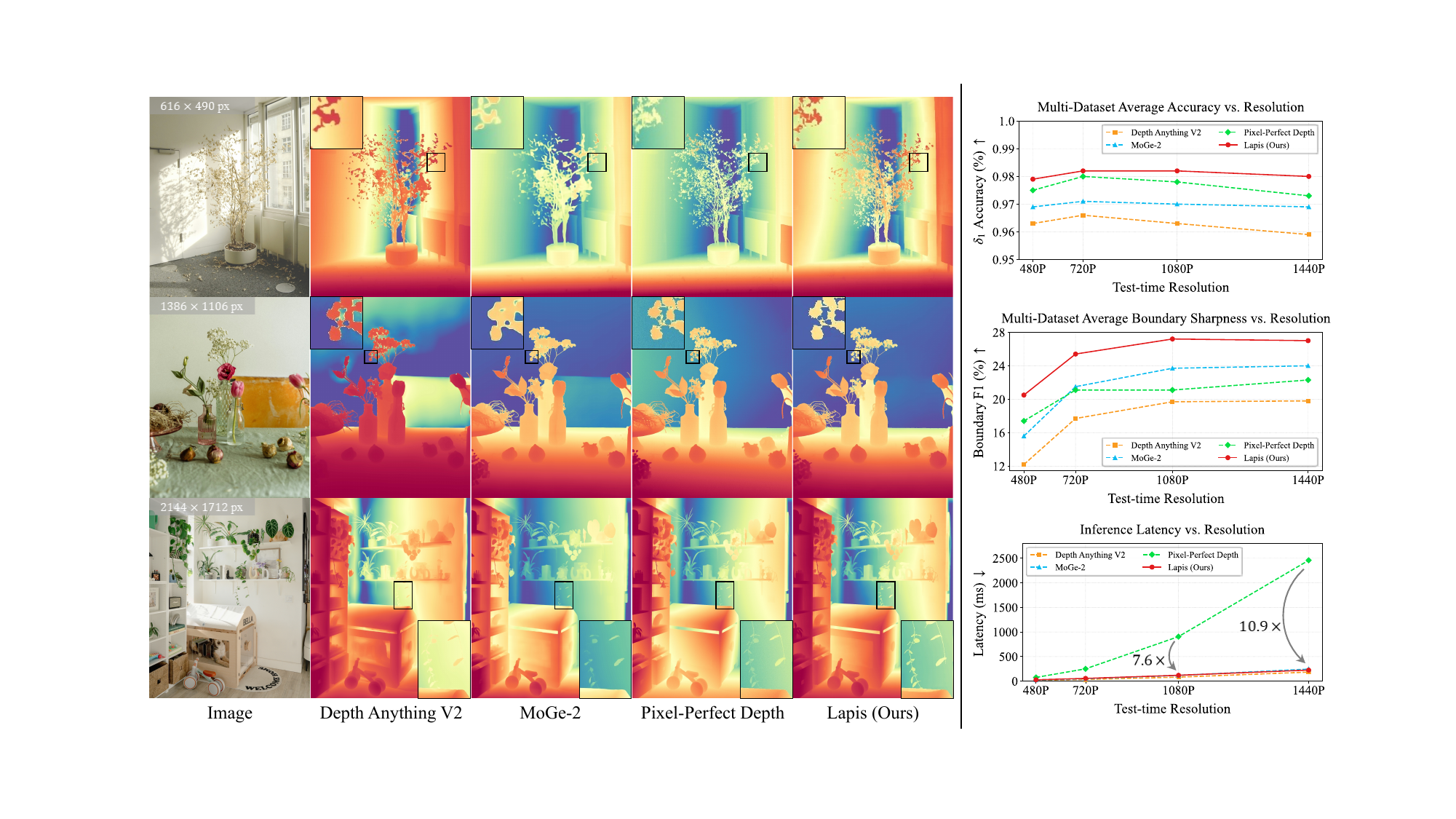}
  \caption{
    \textbf{Lapis: a one-step pixel-space diffusion framework with linear attention for efficient and high-quality monocular depth estimation.}
    Lapis delivers precise geometry and intricate structures across resolutions, consistently outperforming SOTA models in accuracy and boundary sharpness with competitive latency.
  }
  \label{fig:teaser}
\end{figure}

\begin{abstract}
This work presents \textbf{Lapis}, a \textbf{l}inear-\textbf{a}ttention-based \textbf{pi}xel-\textbf{s}pace generative framework that achieves efficient and high-fidelity depth estimation with one-step diffusion.
While generative frameworks have significantly advanced monocular depth estimation with superior detail fidelity, the $\mathcal{O}(N^2)$ complexity of standard attention and the multi-step denoising process introduce prohibitive computational costs when scaling them to high-resolution image applications.
Although linear attention and one-step prediction are intuitively viable, directly applying them leads to poor structural consistency, detail loss, and noise.
Lapis rectifies these limitations through a coarse-to-fine hierarchy. 
Specifically, a Patch-level Consistency Module restores structural coherence by integrating semantic and spatial priors. 
Subsequently, a Pixel-level Refinement Module recovers sharp geometric boundaries via skip-connection-based pixel correspondence.
Furthermore, to mitigate sampling noise inherent in one-step diffusion, we leverage the manifold assumption and adopt a direct $\mathbf{x}$-prediction strategy to target the clean data manifold.
Extensive evaluations on multiple benchmarks demonstrate that Lapis consistently achieves state-of-the-art (SOTA) accuracy and boundary sharpness across various resolutions, reducing inference latency by up to 7.6$\times$ at 1080P and 10.9$\times$ at 1440P resolution compared to previous SOTA generative models.

\end{abstract}

\section{Introduction}

\begin{figure}[tb]
  \centering
  \includegraphics[
    width=\textwidth,
    trim=2.5cm 5.5cm 2.5cm 4cm,
    clip
  ]{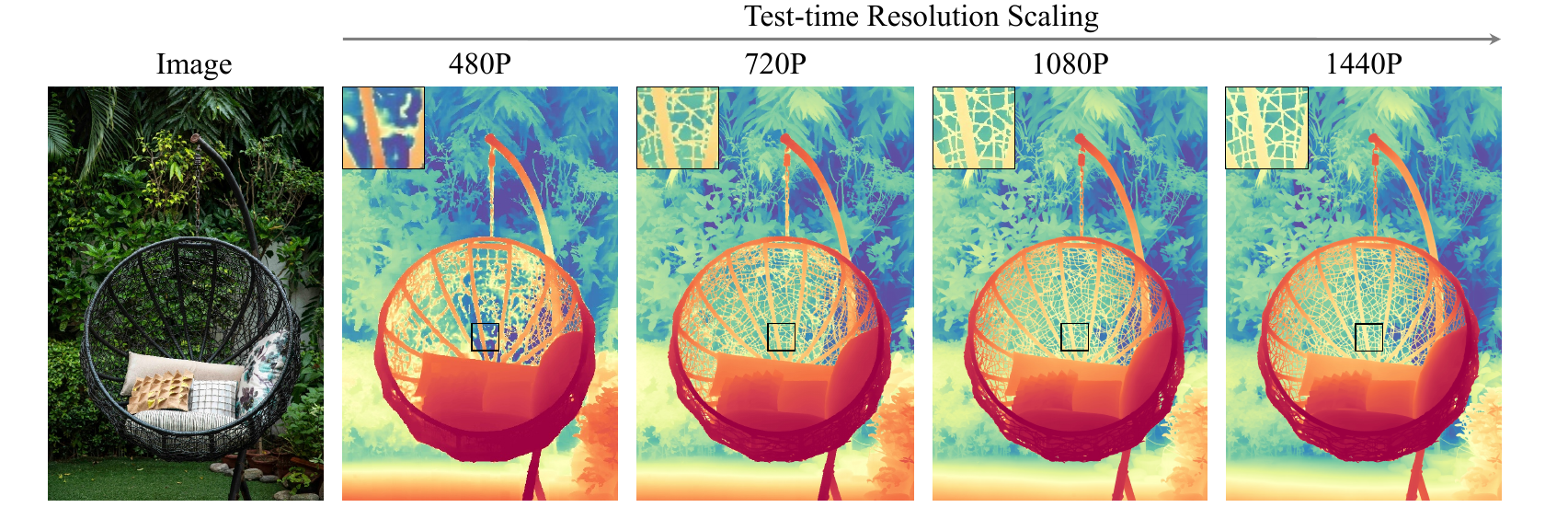}
  \caption{
    \textbf{Visualization of test-time resolution scaling using our model.} 
    A depth estimation model produces more fine-grained details as input resolution increases.
  }
  \label{fig:test-time_resolution_scaling}
\end{figure}

Monocular depth estimation is a cornerstone of computer vision, powering critical applications
in real-world 3D/4D perception and reconstruction \cite{li2023fbocc, Xu_2025_ICCV, dreamvla25, Wu_2025_ICCV, kumar2024few, Li_2024_CVPR, huang2025vivid4dimproving4dreconstruction}.
Driven by the demand for fine-grained geometric perception, the field has recently transitioned from discriminative regression \cite{Ranftl2022, depth_anything_v1, depth_anything_v2, depthanything3, wang2025moge, wang2025moge2} to generative frameworks \cite{ke2023repurposing, ke2025marigold, he2024lotus, gui2024depthfm, fu2024geowizard, xu2024diffusion, martingarcia2024diffusione2eft, xu2025pixel}. 
Compared to their discriminative counterparts, generative models significantly advance fine-grained structural details.
However, as applications increasingly demand higher resolutions to capture intricate, pixel-level geometry, as illustrated in \cref{fig:test-time_resolution_scaling}, existing generative frameworks inevitably face a fundamental trade-off between fidelity and efficiency.
On the one hand, most models \cite{ke2023repurposing, ke2025marigold, he2024lotus, gui2024depthfm, fu2024geowizard, xu2024diffusion, martingarcia2024diffusione2eft} inherit latent-space U-Net \cite{ronneberger2015unetconvolutionalnetworksbiomedical} architectures from pre-trained image generators \cite{rombach2021highresolution} for efficiency.
Such a modeling paradigm is not only restricted by the irreversible information loss of the VAE \cite{kingma2022autoencodingvariationalbayes} bottleneck, but also struggles to capture long-range dependencies necessary for global consistency at high resolutions.
On the other hand, while recent works \cite{xu2025pixel} employ pixel-space Diffusion Transformers (DiTs) \cite{Peebles2022DiT} to eliminate the latent bottleneck and maintain long-range coherence, their $\mathcal{O}(N^2)$ complexity becomes computationally prohibitive at high resolutions.
Moreover, the iterative multi-step sampling in both paradigms multiplies these overheads.
Under these constraints, achieving a trajectory that harmonizes pixel-level precision with inference efficiency remains an open challenge.

In search of an architecture capable of capturing long-range consistency in pixel-space without quadratic complexity or multi-step overhead, an intuitively viable choice is to adopt linearized alternatives \cite{katharopoulos2020transformers} for attention mechanisms \cite{NIPS2017_3f5ee243} in DiTs with a one-step sampling strategy.
Yet, such naive adaptation incurs significant precision loss compared to vanilla DiTs \cite{Peebles2022DiT}.
Specifically, we identify this degradation as stemming from two distinct scales: 
a patch-level structural misalignment that disrupts global consistency, 
and a pixel-level fidelity loss that hampers fine-grained geometric recovery.
Beyond these architectural weaknesses, the widely-adopted $\mathbf{v}$-prediction trajectories \cite{lipman2023flowmatchinggenerativemodeling, liu2022flowstraightfastlearning, albergo2023buildingnormalizingflowsstochastic} exhibit severe instability in pixel-space when restricted to a single forward pass, yielding noise-corrupted results without iterative refinement.

To address these multifaceted challenges, we propose three core innovations.
First, we introduce the Patch-level Consistency Module (PCM) to rectify structural misalignment.
The PCM leverages pre-trained semantic priors \cite{oquab2023dinov2} to enforce global coherence while incorporating local convolutional biases to ensure seamless structural continuity across adjacent patches.
Second, to mitigate pixel-level fidelity loss, we develop the Pixel-level Refinement Module (PRM), which employs skip-connection-based pixel correspondence to recover sharp boundaries and intricate structures.
Third, we stabilize one-step inference by adopting direct pixel-space $\mathbf{x}$-prediction over the volatile $\mathbf{v}$-prediction.
Grounded in the manifold assumption \cite{li2026basicsletdenoisinggenerative}, which posits that natural data resides on a low-dimensional manifold, whereas noise-corrupted quantities do not, this strategy targets the clean data manifold directly, effectively suppressing the sampling instability inherent in one-step generation.
Building on these components, we present \textbf{Lapis}, the first \textbf{l}inear-\textbf{a}ttention-based \textbf{pi}xel-\textbf{s}pace generative framework to enable consistent and high-fidelity depth reconstruction in a single forward pass while maintaining $\mathcal{O}(N)$ computational scaling in the diffusion process.

Experiments across 12 benchmarks demonstrate that Lapis achieves SOTA accuracy and boundary sharpness, consistently surpassing leading discriminative and generative baselines across varying resolutions with competitive efficiency, as shown in \cref{fig:teaser}.
For instance, at 1080P resolution, Lapis maintains a 5.1 average AbsRel (\cref{tab:comp_relative_highres}), a 10\% relative improvement over the strongest competitor, while requiring only 13\% of its latency (\cref{tab:comp_efficiency}).
These results underscore Lapis as an efficient and high-quality solution for high-resolution 3D perception.

Our main contributions are summarized as follows:
\begin{itemize}
    \item We present Lapis, the first linearized pixel-space diffusion model for monocular depth estimation. It enables linear-time depth diffusion within a single forward pass, significantly outperforming generative alternatives in efficiency.
    \item We introduce a coarse-to-fine strategy to rectify the structural and detail degradation of linear attention. This includes a Patch-level Consistency Module that restores global structural coherence by integrating semantic and spatial priors, and a Pixel-level Refinement Module that recovers fine-grained geometric details via skip-connection-based pixel correspondence.
    \item We leverage the manifold assumption to address the unstable one-step pixel-space sampling via a direct $\mathbf{x}$-prediction strategy. This approach effectively suppresses sampling noise and produces clean depth maps.
    \item Extensive evaluations show that Lapis achieves SOTA accuracy and boundary sharpness across multiple benchmarks while maintaining competitive latency, offering a holistic solution for high-resolution 3D perception.
\end{itemize}

\section{Related Work}

\subsection{Generative Monocular Depth Estimation} 

As a cornerstone for 3D vision, monocular depth estimation (MDE) has evolved from early CNN-based regression on specific domains \cite{eigen2014depth, fu2018deep, lee2019big} toward open-domain expansion leveraging diverse datasets \cite{Ranftl2022, Yin2019enforcing}, and finally to the architectural transformation afforded by large-scale Vision Transformers (ViTs) \cite{Ranftl2021, dosovitskiy2021imageworth16x16words, eftekhar2021omnidata, bhat2023zoedepth}.
Recent SOTA discriminative models, such as \cite{depth_anything_v1, depth_anything_v2, depthanything3, wang2025moge, wang2025moge2}, utilize powerful pre-trained encoders \cite{oquab2023dinov2} and massive training data to achieve remarkable zero-shot robustness.
Despite their impressive accuracy and efficiency, these regression-based paradigms often produce over-smoothed predictions that discard fine-grained geometric details.
To overcome this challenge, generative modeling works treat depth estimation as a conditional synthesis task.
One prominent branch \cite{ke2023repurposing, ke2025marigold, he2024lotus, he2025lotus, fu2024geowizard, xu2024diffusion, martingarcia2024diffusione2eft, gui2024depthfm} fine-tunes pre-trained latent-space image synthesis models \cite{rombach2021highresolution, flux2024} to recover intricate structures during the diffusion process.
However, these models remain inadequate for high-resolution depth estimation where high-fidelity geometric details are paramount.
Their U-Net backbones \cite{ronneberger2015unetconvolutionalnetworksbiomedical} struggle with long-range consistency, while VAE latent downsampling \cite{kingma2022autoencodingvariationalbayes} irreversibly discards fine-grained geometric details.
To circumvent these constraints, recent research \cite{xu2025pixel} has shifted toward pixel-space diffusion Transformers (DiTs) \cite{Peebles2022DiT} to preserve complete spatial information for superior detail restoration.
However, pixel-space diffusion predominantly relies on standard attention \cite{NIPS2017_3f5ee243} with $\mathcal{O}(N^2)$ complexity, rendering it computationally prohibitive for high-resolution applications.
Furthermore, iterative diffusion sampling compounds the latency overhead, posing significant challenges for efficient high-resolution deployment.
By contrast, Lapis bridges the gap between geometric precision and inference efficiency via a linearized pixel-space DiT that enables high-fidelity, single-pass reconstruction.
Unlike naive adaptations, our dual-level refinement and $\mathbf{x}$-prediction strategy ensure that these efficiency gains enhance, rather than compromise, the restoration of intricate geometric structures.

\subsection{Sampling Strategies for Generative Depth Estimation}

Early generative MDE frameworks \cite{ke2023repurposing, fu2024geowizard} leverage pre-trained latent diffusion models (LDMs) \cite{rombach2021highresolution} with $\boldsymbol{\epsilon}$-prediction \cite{ho2020denoising}.
Consequently, they inherit the heavy iterative sampling schedule (10--50 steps), leading to substantial computational latency.
To accelerate inference, recent research has branched into two primary directions.
On one hand, several approaches \cite{gui2024depthfm, ke2025marigold, xu2025pixel} employ flow-matching with $\mathbf{v}$-prediction for straighter sampling trajectories, but still require at least 4 steps to sufficiently suppress noise artifacts.
On the other hand, existing one-step $\mathbf{x}$-prediction models \cite{martingarcia2024diffusione2eft, xu2024diffusion, he2024lotus} remain fundamentally constrained by the LDM paradigm.
Specifically, the VAE-based latent space \cite{kingma2022autoencodingvariationalbayes} is explicitly regularized toward an isotropic Gaussian distribution, inherently compromising the low-dimensional manifold of natural geometry \cite{li2026basicsletdenoisinggenerative}.
As a result, performing one-step $\mathbf{x}$-prediction within such a distorted space leads to poor results.
In contrast to existing $\mathbf{v}$-prediction or latent-space $\mathbf{x}$-prediction models, Lapis performs $\mathbf{x}$-prediction directly in pixel-space, thereby preserving the original geometric manifold and achieving SOTA accuracy within a single forward pass.

\subsection{Linear Attention}

To maintain global context while bypassing the quadratic scaling of standard self-attention \cite{NIPS2017_3f5ee243}, linear attention \cite{katharopoulos2020transformers} utilizes kernel approximations to eliminate the Softmax operation, achieving $\mathcal{O}(N)$ complexity.
Despite its widespread adoption in natural language processing \cite{NIPS2017_3f5ee243, zhen2022cosformer, xiong2021nystromformernystrombasedalgorithmapproximating} for long-sequence modeling, it remains disproportionately sparse in visual tasks.
Existing explorations \cite{han2023flatten, Fan_2025_ICCV_magnitude_aware, fan2025breaking, cai2024efficientvitmultiscalelinearattention, lu2022softsoftmaxfreetransformerlinear, han2024demystify} primarily refine the linear attention formulation for categorical tasks with limited data, such as image classification, object detection, instance segmentation, and semantic segmentation, which rely on coarse-grained semantic features and are inherently tolerant of feature fluctuations.
In contrast, MDE is a dense regression task requiring high-precision numerical fidelity, making it far more sensitive to the representational decay often associated with linear approximations.
More recently, generative models \cite{xie2024sana, chen2025sana} have adopted linear attention for visual synthesis tasks.
But in these contexts, visual plausibility, \eg, FID \cite{NIPS2017_8a1d6947} and FVD \cite{unterthiner2019fvd}, often takes precedence over metric accuracy.
Lapis stands as the first application of linear attention for MDE.
We demonstrate that linearized models can achieve high-quality 3D dense perception efficiently without compromising accuracy.

\section{Method}

\begin{figure}[tb]
  \centering
  \includegraphics[
    width=\textwidth,
    trim=4.0cm 0.2cm 4.5cm 0.0cm,
    clip
  ]{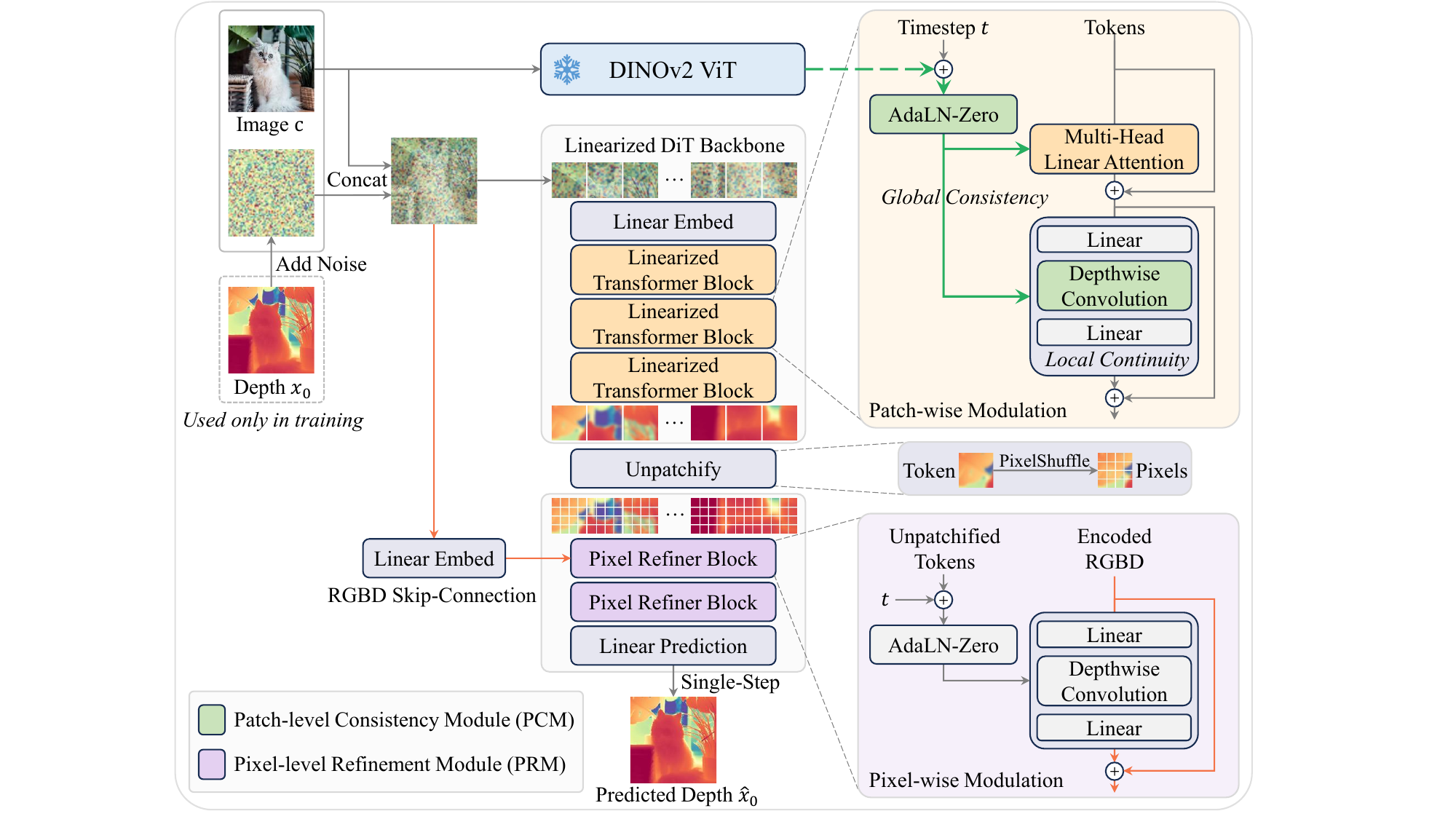}
  \caption{
      \textbf{Architecture overview.}
      Lapis performs pixel-space $\mathbf{x}$-prediction to enable high-quality one-step denoising. 
      It first leverages a linearized DiT backbone to establish a coherent global depth layout, integrating semantic and spatial priors via our Patch-level Consistency Module.
      Subsequently, the Pixel-level Refinement Module restores fine-grained geometry via skip-connection-based pixel correspondence.
      The overall framework enables efficient and accurate depth estimation with pixel-level details.
  }
  \label{fig:pipeline}
\end{figure}

Lapis is designed as a high-efficiency, one-step generative framework for relative monocular depth estimation.
As illustrated in \cref{fig:pipeline}, the model operates directly in pixel-space, taking a concatenated noisy depth map and a conditioning image as input to predict a clean depth map without the need for latent autoencoders.
To circumvent the quadratic complexity inherent in standard DiTs \cite{Peebles2022DiT}, we replace all standard multi-head self-attention (MSA) layers with linearized attention kernels \cite{katharopoulos2020transformers}.
To bridge the representational fidelity gap often associated with linear approximations, we augment the transformer blocks with two synergistic components: the Patch-level Consistency Module (PCM) and the Pixel-level Refinement Module (PRM). 
While the PCM enforces spatial alignment across the token grid at the patch-level, the PRM restores high-frequency precision at the pixel-level, collectively harmonizing linear-time computational efficiency with high-fidelity depth estimation.

\subsection{One-step Pixel-Space Diffusion}

To eliminate the multi-step sampling latency inherent in iterative diffusion models, we adopt a native one-step framework.
Guided by the low-dimensional manifold assumption \cite{li2026basicsletdenoisinggenerative}, which posits that natural geometric data resides on a structured and continuous manifold, our model directly predicts the clean depth map in pixel-space.
Specifically, we formulate the diffusion process following the flow-matching paradigm \cite{lipman2023flowmatchinggenerativemodeling, liu2022flowstraightfastlearning, albergo2023buildingnormalizingflowsstochastic}.
Given a clean depth sample $\mathbf{x}_0$, we sample Gaussian noise $\boldsymbol{\varepsilon} \sim \mathcal{N}(0, 1)$, time step $t \sim [\tau, 1]$ with a small constant $\tau$ (\eg, 0.05), and apply the linear schedule as:
\begin{equation}
    \mathbf{x}_t = (1 - t) \cdot \mathbf{x}_0 + t \cdot \boldsymbol{\varepsilon}.
    \label{eq:linear_schedule}
\end{equation}
Our model $x_\theta(\cdot)$ then predicts $\mathbf{x}_0$ with $\mathbf{x}_t$, $t$ and corresponding image $\mathbf{c}$ as:
\begin{equation}
    \hat{\mathbf{x}}_0 = x_\theta(\mathbf{x}_t, t, \mathbf{c}).
    \label{eq:prediction_target}
\end{equation}

\subsubsection{Training losses.}

Our model is optimized via a joint objective that combines velocity-based flow supervision and multi-scale geometric regularization.
The velocity loss is defined in $\mathbf{v}$-space following flow-matching frameworks \cite{lipman2023flowmatchinggenerativemodeling, liu2022flowstraightfastlearning, albergo2023buildingnormalizingflowsstochastic}, where the target velocity is given by:
\begin{equation}
    \mathbf{v}_t = \frac{d \mathbf{x}_t}{dt} = \boldsymbol{\varepsilon} - \mathbf{x}_0.
    \label{eq:velocity_definition}
\end{equation}
We supervise the network by minimizing the Mean Squared Error (MSE) between the predicted velocity $\hat{\mathbf{v}}_t = (\mathbf{x}_t - \hat{\mathbf{x}}_0) \ / \ t$ and its ground truth:
\begin{equation}
    \mathcal{L}_{\text{velocity}} = \mathbb{E}_{\mathbf{x}_0, \boldsymbol{\varepsilon}, t} \left[ ||\hat{\mathbf{v}}_t - \mathbf{v}_t||_{l_2}^2 \right].
    \label{eq:mse_loss}
\end{equation}
We further incorporate a multi-scale gradient matching loss directly in $\mathbf{x}$-space following \cite{Ranftl2022, depth_anything_v2} to enhance the sharpness of depth maps:
\begin{equation}
    \mathcal{L}_{\text{grad}} = \mathbb{E}_{\mathbf{x}_0, \boldsymbol{\varepsilon}, t} \left[ \sum_{s \in S} w_s \left\lVert \nabla_s \mathbf{x}_0 - \nabla_s \hat{\mathbf{x}}_0 \right\rVert_{l_1} \right],
    \label{eq:gradient_loss}
\end{equation}
where $\nabla_s$ denotes the spatial gradient operator at different resolution scales $S$ and $w_s$ is the scale-specific weight.
The final loss is a weighted combination:
\begin{equation}
    \mathcal{L}_{\text{total}}= \mathcal{L}_{\text{velocity}} + \lambda_{\text{grad}} \mathcal{L}_{\text{grad}},
    \label{eq:total_loss}
\end{equation}
where $\lambda_{\text{grad}}$ is a hyperparameter balancing accuracy and boundary sharpness.

\subsection{Linearized Transformer Architecture} 

\subsubsection{Linear attention formulation.}

In standard self-attention \cite{NIPS2017_3f5ee243} (for single head case), given query, key, and value matrices $Q, K, V \in \mathbb{R}^{N \times d}$, where $N$ is the number of tokens and $d$ is the feature dimension, the attention output is computed as:
\begin{equation}
    \text{Attn}(Q, K, V) = \text{softmax}\left(\frac{QK^T}{\sqrt{d}}\right)V,
    \label{eq:softmax_attention}
\end{equation}
which incurs a quadratic time complexity of $\mathcal{O}(N^2d)$. 
This cost becomes prohibitive for high-resolution input where $N$ is large (\eg, $1024 \times 768$ pixels equal to $3072$ tokens under patch size $16$).
Instead, linear attention \cite{katharopoulos2020transformers} approximates the softmax similarity with a non-negative kernel function $\phi(\cdot)$. 
By leveraging the associativity of matrix multiplication, the computation can be reformulated as:
\begin{equation}
    \text{Attn}_\text{Linear}(Q_i, K, V) = \frac{\phi(Q_i)\sum_{j = 1}^N \phi(K_j)^T V_j}{\phi(Q_i)\sum_{j = 1}^N \phi(K_j)^T} \text{ for each } i \in [1, N].
    \label{eq:linear_attention}
\end{equation}
With pre-computed global context terms $\sum_{j = 1}^N \phi(K_j)^T V_j \in \mathbb{R}^{d \times d}$ and normalization factor $\sum_{j = 1}^N \phi(K_j)^T \in \mathbb{R}^{d \times 1}$, \cref{eq:linear_attention} avoids the explicit construction of the $N \times N$ attention matrix and reduces the time complexity to $\mathcal{O}(Nd^2)$.
We adopt the standard DiT \cite{Peebles2022DiT} as our generative backbone for its structural simplicity and strong scaling properties.
By substituting all MSA layers inside it with \cref{eq:linear_attention}, we effectively reduce its complexity to $\mathcal{O}(N)$.

\subsubsection{The fidelity gap in linearization.} 

Despite its efficiency, we identify a dual-level fidelity gap when applying linear attention to dense depth estimation.
First, the low-rank nature of linear kernels \cite{Fan_2025_ICCV_magnitude_aware, han2023flatten} often leads to patch-level structural misalignment, where geometric coherence is weakened.
Second, the iterative linear averaging inherent in these layers results in a significant loss of pixel-level detail.
To resolve these issues, we introduce the Patch-level Consistency Module (PCM) and the Pixel-Level Refinement Module (PRM), as detailed below.

\begin{figure}[tb]
  \centering
  \vspace{-5pt}
  \includegraphics[
    width=\textwidth,
    trim=1.7cm 5.4cm 2cm 5.2cm,
    clip
  ]{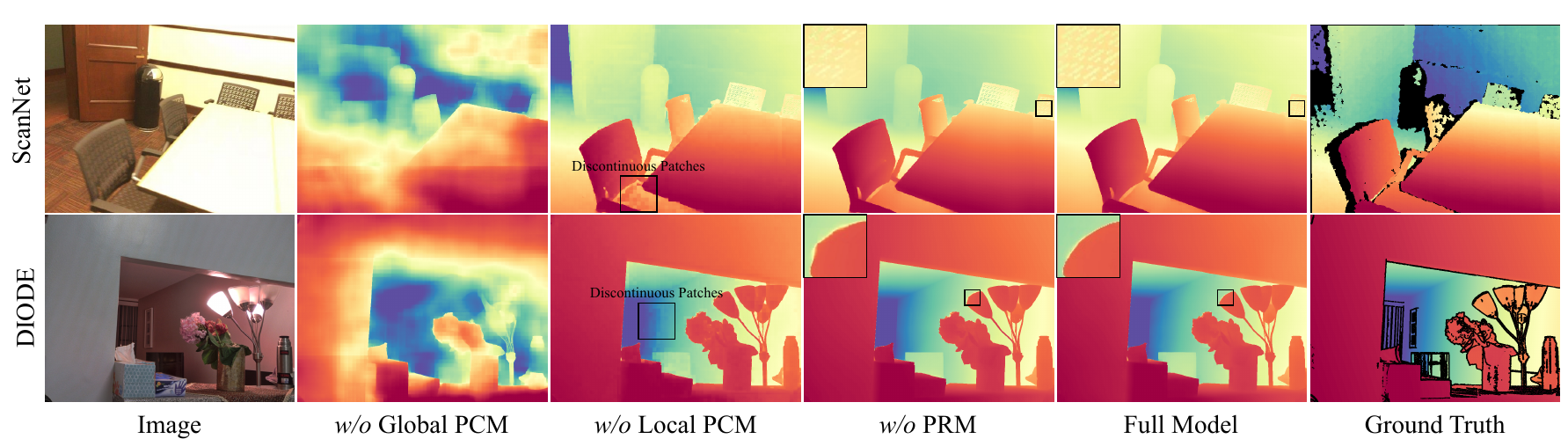}
  \caption{
    \textbf{Qualitative ablation of architectural components.}
    We visually validate the necessity of each model component in Lapis. 
    In comparison with the full model, removing the global PCM leads to severe global inconsistencies, omitting the local PCM introduces discontinuous patch artifacts, and the absence of the PRM results in a significant loss of sharp details.
  }
  \label{fig:component-wise_visualization}
\end{figure}

\subsection{Patch-level Consistency Module}

\subsubsection{Global relational consistency.}

The core limitation of linearized attention lies in the sacrifice of the exponential softmax kernel, which provides a sharp attentional focalization in long-range modeling \cite{han2023flatten, Fan_2025_ICCV_magnitude_aware}.
We observe that forcing linear attention to simultaneously reconstruct global scene layouts and local geometric details is highly impractical.
As illustrated in the second column of \cref{fig:component-wise_visualization}, the model struggles to preserve global geometry layout and subsequently blurs local details.
To alleviate this, we supervise macro-level global geometry with semantic priors, thereby exempting the linearized backbone from the complexities of global spatial coordination.
Specifically, let $\mathbf{s} \in \mathbb{R}^{N \times D}$ denote the DiT tokens, where $N$ and $D$ represent the sequence length and embedding dimension, respectively.
We align $\mathbf{s}$ with semantic descriptors $\mathbf{z} \in \mathbb{R}^{N \times D'}$, which are extracted from the input image $\mathbf{c}$ via a frozen DINOv2 \cite{oquab2023dinov2} encoder.
Here, $D'$ corresponds to the feature dimension of the DINOv2 backbone.
To align and refine high-level semantics, we apply a gating mechanism to $\mathbf{z}$ as follows:
\begin{equation}
    \hat{\mathbf{z}}_i = \mathbf{W}_o \left(\mathbf{W}_{v} \bar{\mathbf{z}}_i \odot \sigma(\mathbf{W}_{g} \bar{\mathbf{z}}_i)\right),
    \label{eq:semantic_alignment}
\end{equation}
where $\{{\mathbf{z}}_i\}_{i = 1}^N$ are tokens normalized from $\mathbf{z}$, $\sigma(\cdot)$ denotes the sigmoid function, $\odot$ represents the element-wise product, and $\mathbf{W}_v, \mathbf{W}_g, \mathbf{W}_o$ are learnable channel-wise projection matrices.
The refined tokens $\hat{\mathbf{z}}$ are subsequently fused with the timestep embedding $\mathbf{t}$ to condition the feature flow in DiT blocks using an extended AdaLN-Zero \cite{Peebles2022DiT} mechanism:
\begin{equation}
    \text{AdaLN-Zero}(f, \mathbf{s}, \mathbf{t}, \hat{\mathbf{z}}) = \boldsymbol{\alpha} \odot f\left( \text{LayerNorm}(\mathbf{s}) \odot (1 + \boldsymbol{\gamma}) + \boldsymbol{\beta} \right),
    \label{eq:adaln_zero}
\end{equation}
where the modulation parameters $(\boldsymbol{\gamma}, \boldsymbol{\beta}, \boldsymbol{\alpha})$ are regressed from the joint embedding $\hat{\mathbf{z}} + \mathbf{t}$ via a multi-layer perceptron (MLP).
We apply such a mechanism to both the linear attention layers and the feed-forward network (FFN) layers  \cite{NIPS2017_3f5ee243}, which are represented by $f(\cdot)$.
In this way, we effectively anchor the feature flow within a consistent global geometric layout without introducing additional parameters to the modulation process.
These semantically-modulated blocks are interleaved in the linearized DiT backbone for computational efficiency.

\subsubsection{Local spatial continuity.}

Despite the global structural rectification, linear attention still struggles to maintain local patch-wise continuity.
As illustrated in the third column of \cref{fig:component-wise_visualization}, this deficiency manifests as localized depth discontinuities and perceptible grid artifacts.
We address this structural fragmentation by incorporating a local inductive bias into the FFN layer via a residual $3 \times 3$ depth-wise convolution (DWC), a structure we term ConvFFN.
Specifically, we formulate the feature transition in the ConvFFN as follows:
\begin{equation}
    \mathbf{s}' = \text{SiLU}(\text{Linear}(\mathbf{s})), \quad \mathbf{s}_{\text{out}} = \text{Linear}(\text{SiLU}(\text{DWC}(\mathbf{s}')) + \mathbf{s}'),
    \label{eq:convffn}
\end{equation}
where $\text{SiLU}(\cdot)$ denotes the SiLU activation function \cite{elfwing2017sigmoidweightedlinearunitsneural}.
This local inductive bias enforces spatial correlation across adjacent tokens, effectively suppressing the aforementioned grid artifacts to ensure smooth geometric transitions.

\begin{figure}[tb]
  \centering
  \includegraphics[
    width=\textwidth,
    trim=1.7cm 1.7cm 1.5cm 1.5cm,
    clip
  ]{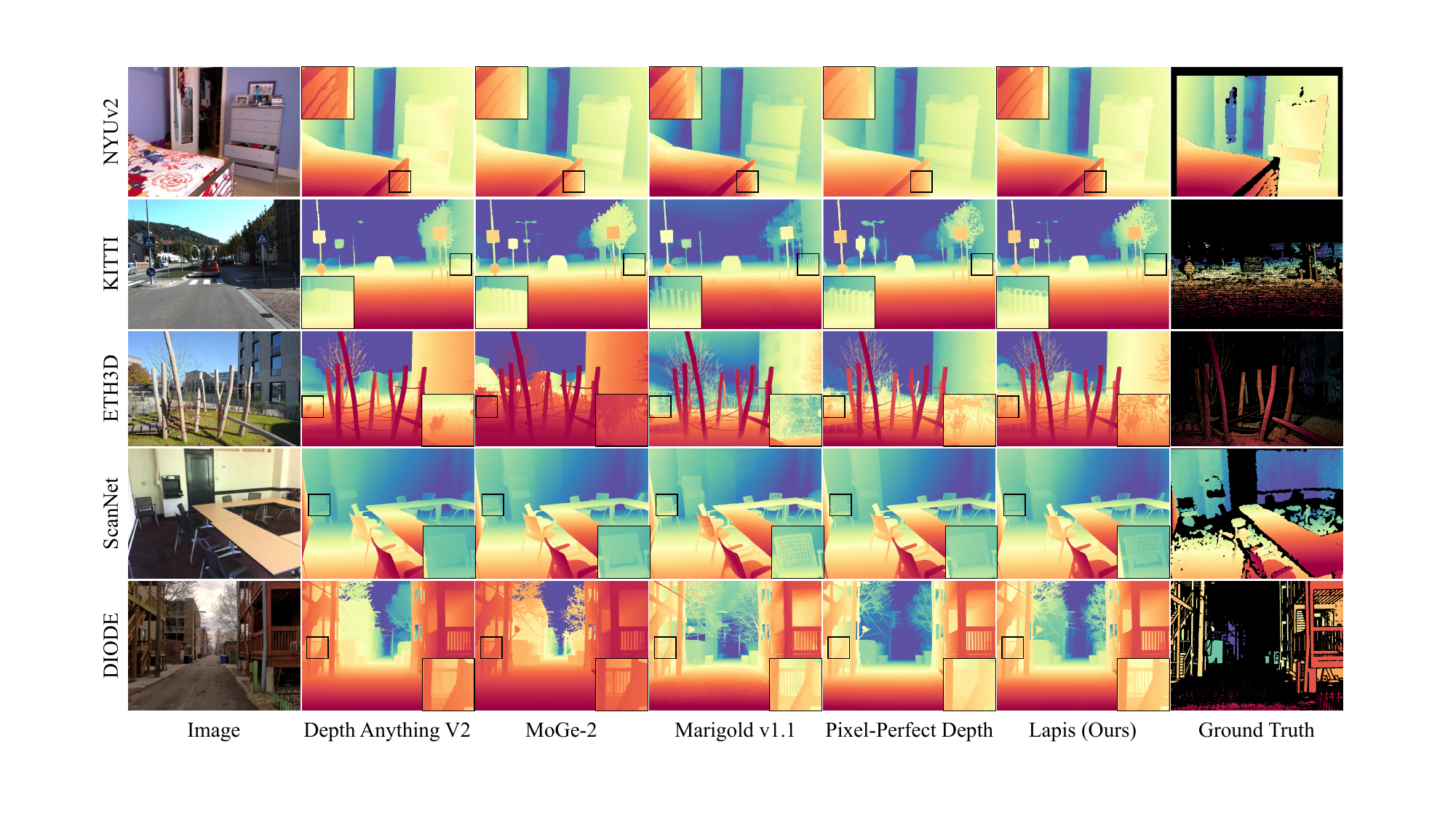}
  \caption{
    \textbf{Qualitative comparison of monocular depth estimation models.}
    Our model excels at capturing thin structures while preserving the accurate global geometry.
    }
  \label{fig:zero-shot_relative_visualization}
\end{figure}

\subsection{Pixel-Level Refinement Module}

Although the PCM ensures token-level coherence, mapping discretized tokens back to the continuous pixel-space still entails a significant fidelity gap.
As previously analyzed, the iterative averaging effect in deep linear-attention-based architectures leads to cumulative signal dilution, where detailed structures progressively fade across layers.
Such degradation is evident in the fourth column of \cref{fig:component-wise_visualization}, where a backbone without PRM fails to resolve sharp geometric boundaries.
To counteract this, the Pixel-level Refinement Module (PRM) re-integrates pixel-level input cues via a long-range skip connection to effectively recover sharp geometric boundaries.
Specifically, the original input $y$ is concatenated with a learnable patch-wise positional encoding and transformed by an MLP to yield the PRM input $y' \in \mathbb{R}^{D_{\text{pixel}} \times H \times W}$, where $D_{\text{pixel}}$ denotes the embedding dimension of the PRM.
To bridge the gap between patch-level representation and pixel-level representation, latent tokens $\mathbf{s} \in \mathbb{R}^{N \times D}$ from the final DiT block are first reshaped into a spatial grid $\mathbb{R}^{D \times h \times w}$ and then projected and upsampled via PixelShuffle \cite{shi2016realtimesingleimagevideo} to match the target pixel-level resolution, following the transformation chain as follows:
\begin{equation}
    \mathbf{s} \xrightarrow[]{\text{MLP}} \mathbf{s}' \in \mathbb{R}^{(D_{\text{pixel}} \times p^2)\times h \times w} \xrightarrow[]{\text{PixelShuffle}} \mathbf{s}_{\text{pixel}} \in \mathbb{R}^{D_{\text{pixel}} \times H \times W},
    \label{eq:pixel_shuffle}
\end{equation}
where $p$ denotes the patch size, and $H = h \cdot p$ and $W = w \cdot p$ represent the height and width of the original input image.
The PRM is architected as a sequence of Pixel Refiner blocks, each using an AdaLN-Zero-modulated ConvFFN layer to decode the pixel-level latents into depth values with pixel-level cues:
\begin{equation}
    \text{AdaLN-Zero}(g, \mathbf{y}', \mathbf{s}_{\text{pixel}}) = \boldsymbol{\alpha} \odot g\left( \text{LayerNorm}(\mathbf{y'}) \odot (1 + \boldsymbol{\gamma}) + \boldsymbol{\beta} \right),
    \label{eq:pixel_refiner}
\end{equation}
where $g(\cdot)$ denotes the ConvFFN layer, and the scaling and shifting parameters $(\boldsymbol{\gamma}, \boldsymbol{\beta}, \boldsymbol{\alpha})$ are regressed from the features $\mathbf{s}_{\text{pixel}}$ via an MLP.
This design effectively recovers pixel-level details from the raw input pixels during the decoding stage.
Finally, a standard linear prediction head in DiT \cite{Peebles2022DiT} is used to project the refined pixel-level features into a high-fidelity depth map $\mathbf{d} \in \mathbb{R}^{H \times W}$.

\section{Experiments}

\begin{table}[t]
  \caption{
      \textbf{Zero-shot relative depth estimation on native resolution.}
      Lapis surpasses both established discriminative models and recent generative baselines.
  }
  \label{tab:comp_relative_standard}
  \centering
  \resizebox{\linewidth}{!}{%
    \begin{tabular}{c c c  cc  cc  cc  cc  cc  cc}
    \toprule
    \multirow{2}{*}{\textbf{Type}} & \multirow{2}{*}{\textbf{Method}} & \multirow{2}{*}{\makecell{\textbf{Training}\\\textbf{Data}}} & \multicolumn{2}{c}{\textbf{NYUv2}} & \multicolumn{2}{c}{\textbf{KITTI}} & \multicolumn{2}{c}{\textbf{ETH3D}} & \multicolumn{2}{c}{\textbf{ScanNet}} & \multicolumn{2}{c}{\textbf{DIODE}} & \multicolumn{2}{c}{\textbf{Average}} \\
    & & & AbsRel$\downarrow$ & $\ \ \delta_1\uparrow \ \ $ & AbsRel$\downarrow$ & $\ \ \delta_1\uparrow \ \ $ & AbsRel$\downarrow$ & $\ \ \delta_1\uparrow \ \ $ & AbsRel$\downarrow$ & $\ \ \delta_1\uparrow \ \ $ & AbsRel$\downarrow$ & $\ \ \delta_1\uparrow \ \ $ & AbsRel$\downarrow$ & $\ \ \delta_1\uparrow \ \ $ \\
    \midrule
    
    \multirow{11}{*}{\rotatebox{90}{\large Discriminative}} 
    & MiDaS \cite{Ranftl2022}
        & 2M     & 11.1 & 88.5 & 23.6 & 63.0 & 18.4 & 75.2 & 12.1 & 84.6 & --   & --   & --   & --\\
    & LeRes \cite{Wei2021CVPR}
        & 354K   & 9.0  & 91.6 & 14.9 & 78.4 & 17.1 & 77.7 & 9.1  & 91.7 & --   & --   & --   & --\\
    & Omnidata \cite{eftekhar2021omnidata}
        & 12.2M  & 7.4  & 94.5 & 14.9 & 83.5 & 16.6 & 77.8 & 7.5  & 93.6 & --   & --   & --   & --\\
    & DPT \cite{Ranftl2021}
        & 1.4M   & 9.8  & 90.3 & 10.0 & 90.1 & 7.8  & 94.6 & 8.2  & 93.4 & --   & --   & --   & --\\
    & HDN \cite{zhang2022hierarchical}         
        & 300K   & 6.9  & 94.8 & 11.5 & 86.7 & 12.1 & 83.3 & 8.0  & 93.9 & --   & --   & --   & --\\
    & Lotus-D \cite{he2024lotus}
        & 59K    & 5.1  & 97.2 & 8.1  & 93.1 & 6.1  & 97.0 & 5.5  & 96.5 & --   & --   & --   & --\\
    & Depth Anything \cite{depth_anything_v1} *
        & 62.6M  & 4.5  & 97.9 & 8.7  & 93.6 & 6.6  & 98.1 & 4.4  & 98.0 & 11.1 & 94.8 & 7.0  & 96.5  \\
    & Depth Anything V2 \cite{depth_anything_v2} * 
        & 62.6M  & 4.5  & 97.9 & 7.9  & 93.6 & 7.4  & 97.7 & 4.2  & 97.8 & 10.0 & 94.9 & 6.8  & 96.4  \\
    & Depth Anything 3 \cite{depthanything3} *
        & 1.2M (Scenes) & 4.1  & 97.4 & 7.0  & 95.6 & 4.2  & 97.7 & 4.1  & 97.6 & 7.1  & 94.6 & 5.3  & 96.6  \\
    & MoGe \cite{wang2025moge} *
        & 9M     & \cellranksecond{3.7}  & 97.9 & \cellrankthird{5.9}  & \cellranksecond{97.3} & \cellrankthird{3.5}  & \cellrankthird{98.4} & \cellranksecond{3.6}  & \cellrankfirst{98.3} & 7.7  & 94.8 & 4.9  & \cellrankthird{97.3}  \\
    & MoGe-2 \cite{wang2025moge2} *
        & 8.9M   & \cellrankfirst{3.5}  & \cellrankthird{98.0} & \cellrankthird{5.9}  & \cellrankthird{97.1} & 4.5  & 96.7 & \cellrankfirst{3.4}  & \cellrankfirst{98.3} & \cellrankthird{6.7}  & \cellrankthird{95.5} & \cellrankthird{4.8}  & 97.1  \\
    \midrule
    
    \multirow{9}{*}{\rotatebox{90}{\large Generative}}
    & GeoWizard \cite{fu2024geowizard}             
        & 280K   & 5.6  & 96.3 & 14.4 & 82.0 & 6.6  & 95.8 & 6.4  & 95.0 & --   & --   & --   & --    \\
    & GenPercept \cite{xu2024diffusion}                
        & 74K    & 5.6  & 96.0 & 13.0 & 84.2 & 7.0  & 95.6 & 6.2  & 96.1 & --   & --   & --   & --    \\
    & Diffusion-E2E-FT \cite{martingarcia2024diffusione2eft}       
        & 74K    & 5.4  & 96.5 & 9.6  & 92.1 & 6.4  & 95.9 & 5.8  & 96.5 & --   & --   & --   & --    \\
    & Lotus-G \cite{he2024lotus}
        & 59K    & 5.4  & 96.8 & 8.5  & 92.2 & 5.9  & 97.0 & 5.9  & 95.7 & --   & --   & --   & --    \\
    & Marigold v1.0 \cite{ke2023repurposing}
        & 74K    & 5.5  & 96.4 & 9.9  & 91.6 & 6.5  & 95.9 & 6.4  & 95.2 & 11.1*& 89.5*& 7.9 & 93.7   \\
    & Marigold v1.1 \cite{ke2025marigold}
        & 74K    & 5.5  & 96.4 & 10.5 & 90.2 & 6.9  & 95.7 & 5.8  & 96.3 & 10.1*& 91.0*& 7.7  & 93.9  \\
    & Lotus-2 \cite{he2025lotus}
        & 59K    & 4.1  & 97.6 & 6.7  & 94.5 & 4.6  & 98.1 & 4.2  & 97.6 & 8.4* & 94.7*& 5.6  & 96.5  \\
    & Pixel-Perfect Depth \cite{xu2025pixel} *
        & 125K   & \cellranksecond{3.7}  & \cellranksecond{98.1} & \cellranksecond{5.4}  & \cellrankthird{97.1} & \cellrankfirst{3.1}  & \cellrankfirst{99.2} & \cellrankthird{3.8}  & \cellrankthird{98.2} & \cellranksecond{4.9}  & \cellranksecond{97.5} & \cellranksecond{4.2}  & \cellranksecond{98.0}  \\  
    & \textbf{Lapis (Ours)} *
        & 122K   & 3.8  & \cellrankfirst{98.2} & \cellrankfirst{5.1}  & \cellrankfirst{97.7} & \cellranksecond{3.2}  & \cellrankfirst{99.2} & \cellrankthird{3.8}  & \cellrankthird{98.2} & \cellrankfirst{4.8}  & \cellrankfirst{97.9} & \cellrankfirst{4.1}  & \cellrankfirst{98.2}  \\  
    \bottomrule
    \end{tabular}%
  }
\end{table}

\subsection{Implementation Details}

\subsubsection{Depth normalization.}

Given raw depth map $\mathbf{d}$, our model follows \cite{xu2025pixel} to predict the normalized log-scale depth $\mathbf{x}_0$ as:
\begin{equation}
    \mathbf{x}_0 = \frac{\hat{\mathbf{d}} - \hat{\mathbf{d}}_{2}}{\hat{\mathbf{d}}_{98} - \hat{\mathbf{d}}_{2}} - 0.5, \label{eq:depth_normalization}
\end{equation}
where $\hat{\mathbf{d}} = \log(\mathbf{d} + \epsilon)$ with the offset $\epsilon$ fixed at $0.01$ to ensure numerical stability. 
$\hat{\mathbf{d}}_{2}$ and $\hat{\mathbf{d}}_{98}$ represent the $2^{\text{nd}}$ and $98^{\text{th}}$ percentiles computed independently for each depth map. 
We further rescale $\mathbf{x}_0$ to ensure unit variance, facilitating a more stable data distribution.

\subsubsection{Training datasets.}

We adopt a total of about 122K training samples from five synthetic datasets: Hypersim \cite{roberts:2021} (53.8K), TartanAir \cite{wang2020tartanair} (28.2K), VKITTI2 \cite{cabon2020vkitti2} (21.3K), UrbanSyn \cite{GOMEZ2025130038} (7.5K), and MatrixCity \cite{li2023matrixcity} (11.6K) to train our model.
All training datasets are directly resized to $1024 \times 768$ resolution and simply concatenated together without weighted sampling. 

\subsubsection{Model architecture details.}

Our model comprises 520M parameters, maintaining consistency with the DiT-Large \cite{Peebles2022DiT} configuration in hidden dimensions and the number of blocks while transitioning to a fully linearized backbone.
The kernel feature map $\phi(\cdot)$ in \cref{eq:linear_attention} is implemented as $\phi(x) = \text{elu}(x) + 1$ \cite{clevert2016fastaccuratedeepnetwork_elu}.
Following \cite{xu2025pixel}, we employ the DINOv2-Large \cite{oquab2023dinov2} encoder fine-tuned by MoGe-2 \cite{wang2025moge2} as the frozen semantic encoder to provide robust structural priors.
For the PRM, we use $M = 3$ blocks and an embedding dimension of $D_{\text{pixel}} = 32$.

\subsubsection{Training details.}

We optimize the model using the total loss defined in \cref{eq:total_loss}, with the gradient loss weight set to $\lambda_{\text{grad}} = 1.5$.
Training is conducted using the AdamW optimizer \cite{loshchilov2019decoupledweightdecayregularization} with a constant learning rate of $1 \times 10^{-4}$ and no weight decay.
The model is trained for 400K steps on 4 NVIDIA RTX Pro 6000 Blackwell GPUs, with a per-GPU batch size of 8.
An Exponential Moving Average (EMA) with a decay rate of $0.9999$ is applied to the model weights during training, and all evaluations are performed using the EMA weights.
Both training and inference are conducted in bfloat16 mixed precision.

\subsection{Zero-shot Relative Depth Evaluation}

\subsubsection{Native resolution.} 

We evaluate zero-shot relative depth estimation on five mainstream unseen real-world datasets: NYUv2 \cite{Silberman:ECCV12} (indoor), KITTI \cite{Geiger2013IJRR} (outdoor), ETH3D \cite{schoeps2017cvpr} (various), ScanNet \cite{dai2017scannet} (indoor), and DIODE \cite{diode_dataset} (various). 
We adopt two widely accepted metrics: Absolute Relative Error (AbsRel) and $\delta_1$ accuracy.
\footnote{
    *indicates results re-evaluated using the evaluation protocol of Marigold \cite{ke2023repurposing, ke2025marigold}. 
}
Each model is evaluated at its training resolution (usually lower than 720P) to assess its peak accuracy.
As summarized in \cref{tab:comp_relative_standard}, Lapis achieves the best performance with an averaged AbsRel of $4.1\%$ and a $\delta_1$ accuracy of $98.2\%$.
Besides, as demonstrated in \cref{fig:zero-shot_relative_visualization}, our model excels in resolving thin structures and intricate details while maintaining a coherent global geometric layout.

\subsubsection{Test-time resolution scaling at 1080P and 1440P.}

\begin{table}[t]
  \caption{
      \textbf{Zero-shot relative depth estimation on 1080P and 1440P.} 
      Notably, Lapis demonstrates the exceptional resolution-robustness of our dual-level PCM and PRM modeling and largely outperforms existing baselines in accuracy metrics.
  }
  \label{tab:comp_relative_highres}
  \centering
  \resizebox{\linewidth}{!}{%
    \begin{tabular}{c c cc cc cc cc cc cc}
    \toprule
    \multirow{2}{*}{\textbf{Res.}} & \multirow{2}{*}{\textbf{Method}} & 
    \multicolumn{2}{c}{\makecell{\textbf{Middlebury} \\ $1920 \times 1080$}} & 
    \multicolumn{2}{c}{\makecell{\textbf{Booster} \\ $4112 \times 3008$}} & 
    \multicolumn{2}{c}{\makecell{\textbf{Cityscapes} \\ $2048 \times 1024$}} & 
    \multicolumn{2}{c}{\makecell{\textbf{DrivingStereo} \\ $1762 \times 800$}} & 
    \multicolumn{2}{c}{\makecell{\textbf{ETH3D} \\ $6048 \times 4032$}} &
    \multicolumn{2}{c}{\makecell{\textbf{Average} \\ -- --}} \\
    & & \small AbsRel$\downarrow$ & \small \ \ $\delta_1\uparrow$ \ \ & \small AbsRel$\downarrow$ & \small \ \ $\delta_1\uparrow$ \ \ & \small AbsRel$\downarrow$ & \small \ \ $\delta_1\uparrow$ \ \ & \small AbsRel$\downarrow$ & \small \ \ $\delta_1\uparrow$ \ \ & \small AbsRel$\downarrow$ & \small \ \ $\delta_1\uparrow$ \ \ & \small AbsRel$\downarrow$ & \ \ $\delta_1\uparrow$ \ \ \\
    \midrule

    \multirow{7}{*}{\rotatebox{90}{1080P}} 
    & Depth Anything V2 \cite{depth_anything_v2}      
        & \cellrankthird{6.6}  & \cellrankfirst{96.8} & 3.3  & 99.7 & \cellrankthird{10.1} & \cellrankthird{89.4} & 6.4 & 96.4 & 5.7  & \cellrankthird{98.1} & 6.4  & \cellrankthird{96.1} \\
    & Depth Anything 3 \cite{depthanything3}
        & 7.6  & 93.4 & \cellrankthird{2.4}  & \cellrankthird{99.8} & 18.9 & 75.3 & \cellranksecond{5.7} & \cellranksecond{96.8} & 5.1  & 97.5 & 7.9  & 92.6 \\
    & MoGe-2 \cite{wang2025moge2}
        & 7.5  & 93.8 & \cellrankfirst{2.0}  & \cellrankfirst{99.9} & 11.8 & 89.2 & \cellrankthird{5.8} & 96.5 & \cellrankthird{4.5}  & 96.9 & \cellrankthird{6.3}  & 95.3 \\
    & Marigold v1.1 \cite{ke2025marigold}
        & 13.8 & 81.4 & 8.6  & 94.3 & 24.0 & 60.0 & 7.1 & 95.9 & 11.5 & 87.2 & 13.0 & 83.8 \\
    & Lotus-2 \cite{he2025lotus}
        & 14.5 & 76.2 & 8.0  & 91.3 & 31.8 & 58.1 & 6.1 & 96.6 & 6.6  & 97.5 & 13.4 & 83.9 \\
    & Pixel-Perfect Depth \cite{xu2025pixel}
        & \cellranksecond{6.3}  & \cellrankthird{96.5} & 2.9  & \cellrankthird{99.8} & \cellranksecond{9.3}  & \cellrankfirst{93.5} & 6.1 & \cellranksecond{96.8} & \cellranksecond{4.1}  & \cellranksecond{99.1} & \cellranksecond{5.7}  & \cellrankfirst{97.1} \\
    & \textbf{Lapis (Ours)}     
        & \cellrankfirst{5.6}  & \cellrankfirst{96.8} & \cellranksecond{2.1}  & \cellrankfirst{99.9} & \cellrankfirst{8.7}  & \cellranksecond{92.5} & \cellrankfirst{5.6} & \cellrankfirst{97.0} & \cellrankfirst{3.3}  & \cellrankfirst{99.3} & \cellrankfirst{5.1}  & \cellrankfirst{97.1} \\
    \midrule
    
    \multirow{7}{*}{\rotatebox{90}{1440P}}
    & Depth Anything V2 \cite{depth_anything_v2}   
        & \cellranksecond{6.9}  & \cellrankfirst{96.5} & 3.4  & 99.6 & \cellrankthird{10.6} & 88.7 & 6.4 & 96.5 & 5.7  & \cellrankthird{97.9} & 6.6  & \cellrankthird{95.8} \\
    & Depth Anything 3 \cite{depthanything3} 
        & 7.9  & 93.4 & \cellrankthird{2.9}  & \cellrankfirst{99.9} & 20.4 & 71.4 & \cellrankfirst{5.6} & \cellranksecond{96.9} & 6.2  & 96.6 & 8.6  & 91.6 \\
    & MoGe-2 \cite{wang2025moge2}
        & 7.7  & 93.5 & \cellrankfirst{2.0}  & \cellrankfirst{99.9} & 12.0 & \cellrankthird{88.9} & \cellrankthird{5.8} & 96.5 & \cellranksecond{4.6}  & 96.8 & \cellranksecond{6.4}  & 95.1 \\
    & Marigold v1.1 \cite{ke2025marigold}
        & 17.1 & 73.9 & 10.4 & 91.2 & 40.8 & 42.1 & 7.3 & 95.3 & 15.8 & 78.6 & 18.3 & 76.2 \\
    & Lotus-2 \cite{he2025lotus}
        & 16.6 & 71.1 & 10.0 & 88.9 & 65.1 & 47.5 & 6.2 & 96.6 & 8.4  & 96.3 & 21.3 & 80.1 \\
    & Pixel-Perfect Depth \cite{xu2025pixel}
        & \cellrankthird{7.2}  & \cellrankthird{95.3} & 3.9  & 99.8 & \cellranksecond{10.0} & \cellrankfirst{92.9} & 6.5 & \cellrankthird{96.7} & \cellrankthird{5.1}  & \cellranksecond{98.5} & \cellrankthird{6.5}  & \cellranksecond{96.6} \\
    & \textbf{Lapis (Ours)}     
        & \cellrankfirst{5.8}  & \cellranksecond{96.4} & \cellranksecond{2.1}  & \cellrankfirst{99.9} & \cellrankfirst{9.1}  & \cellranksecond{91.8} & \cellranksecond{5.7} & \cellrankfirst{97.0} & \cellrankfirst{3.4}  & \cellrankfirst{99.2} & \cellrankfirst{5.2}  & \cellrankfirst{96.9} \\
    \bottomrule
    \end{tabular}%
  }
\end{table}

Benefiting from linearized attention and one-step diffusion, Lapis enables efficient scaling to high resolutions for enhanced geometric details.
To validate that such a capability does not come at the cost of degraded global accuracy, we evaluate zero-shot relative depth at 1080P and 1440P test-time resolution on five datasets with high-resolution annotations: Middlebury \cite{scharstein2014high} (indoor), Booster \cite{zamaramirez2022booster, zamaramirez2024booster} (indoor), Cityscapes \cite{Cordts2016Cityscapes} (outdoor), DrivingStereo \cite{yang2019drivingstereo} (outdoor), and ETH3D \cite{schoeps2017cvpr} (various).
As reported in \cref{tab:comp_relative_highres}, Lapis maintains remarkable accuracy metrics, outperforming current SOTA models by approximately $10\%$ and $19\%$ in AbsRel, respectively.
This demonstrates that our architecture not only scales efficiently but also preserves global structural integrity under significant resolution shifts.

\subsection{Boundary Sharpness Evaluation}

\begin{table}[t]
  \caption{
      \textbf{Boundary sharpness across different resolutions.} 
      We report the F1-score ($\uparrow$) for depth boundary accuracy.
      Lapis consistently outperforms all existing methods across various resolutions, while achieving near-monotonic improvements in boundary sharpness as the test-time resolution scales up.
  }
  \label{tab:comp_boundary_sharpness}
  \centering
  \resizebox{\linewidth}{!}{%
    \begin{tabular}{c cccc cccc cccc cccc}
    \toprule
    \multirow{2}{*}{\textbf{Method}} & \multicolumn{4}{c}{\textbf{Hypersim F1$\uparrow$}} & \multicolumn{4}{c}{\textbf{Sintel F1$\uparrow$}} & \multicolumn{4}{c}{\textbf{Spring F1$\uparrow$}} & \multicolumn{4}{c}{\textbf{Average F1$\uparrow$}}\\
    & 480P & 720P & 1080P & 1440P 
    & 480P & 720P & 1080P & 1440P 
    & 480P & 720P & 1080P & 1440P 
    & 480P & 720P & 1080P & 1440P \\
    \midrule

    Depth Anything V2 \cite{depth_anything_v2} &
        13.1 & 20.9 & 24.5 & 24.1 & 17.3 & \cellrankthird{23.0} & \cellrankthird{23.7} & 23.6 & \cellrankthird{6.1}  & \cellrankthird{9.2}  & \cellrankthird{11.0} & \cellrankthird{11.7} & 12.2 & 17.7 & 19.7 & 19.8 \\
    Depth Anything 3 \cite{depthanything3} &
        12.5 & 19.3 & 23.1 & 24.4 & 15.8 & 20.9 & \cellrankthird{23.7} & \cellrankthird{24.6} & 5.8  & 8.7  & 10.2 & 10.3 & 11.4 & 16.3 & 19.0 & 19.8 \\
    MoGe-2 \cite{wang2025moge2} &
        19.5 & \cellrankthird{28.4} & \cellrankthird{31.7} & \cellranksecond{32.1} & \cellranksecond{21.4} & \cellranksecond{27.8} & \cellranksecond{29.2} & \cellranksecond{28.9} & 5.8  & 8.4  & 10.3 & 10.9 & \cellrankthird{15.6} & \cellranksecond{21.5} & \cellranksecond{23.7} & \cellranksecond{24.0} \\
    Marigold v1.1 \cite{ke2025marigold} &
        \cellrankthird{20.9} & 22.2 & 8.9  & 7.7  & 17.5 & 18.8 & 6.1  & 5.6  & 5.0  & 5.6  & 2.9  & 2.6  & 14.5 & 15.6 & 6.0  & 5.3  \\
    Lotus-2 \cite{he2025lotus} &
        17.1 & 21.9 & 22.6 & 22.0 & 14.5 & 18.3 & 19.2 & 18.5 & 4.7  & 7.5  & 9.3  & 9.4  & 12.1 & 15.9 & 17.0 & 16.6 \\
    Pixel-Perfect Depth \cite{xu2025pixel} &
        \cellranksecond{26.5} & \cellranksecond{31.2} & \cellranksecond{32.7} & \cellrankthird{30.6} & \cellrankthird{19.1} & 21.5 & 21.0 & 20.8 & \cellranksecond{6.5}  & \cellranksecond{10.6} & \cellranksecond{12.0} & \cellranksecond{13.2} & \cellranksecond{17.4} & \cellrankthird{21.1} & \cellrankthird{21.1} & \cellrankthird{22.3} \\
    \textbf{Lapis (Ours)} &
        \cellrankfirst{27.4} & \cellrankfirst{32.2} & \cellrankfirst{34.8} & \cellrankfirst{33.6} & \cellrankfirst{26.2} & \cellrankfirst{30.9} & \cellrankfirst{31.4} & \cellrankfirst{31.7} & \cellrankfirst{8.0}  & \cellrankfirst{13.0} & \cellrankfirst{15.5} & \cellrankfirst{15.8} & \cellrankfirst{20.5} & \cellrankfirst{25.4} & \cellrankfirst{27.2} & \cellrankfirst{27.0} \\
    \bottomrule
    \end{tabular}%
  }
\end{table}
\begin{table}[t]
  \caption{
      \textbf{Runtime statistics across different resolutions.}
      We report latency (ms), throughput (Hz), and peak GPU memory (GB).
      Lapis achieves throughput competitive with leading discriminative models while consistently outperforming existing generative frameworks by a significant margin.
  }
  \label{tab:comp_efficiency}
  \centering
  \resizebox{\linewidth}{!}{%
    \begin{tabular}{c @{\hspace{12pt}} ccc ccc ccc ccc}
    \toprule
    \multirow{2}{*}{\textbf{Method}} & \multicolumn{3}{c}{\textbf{480P}} & \multicolumn{3}{c}{\textbf{720P}} & \multicolumn{3}{c}{\textbf{1080P}} & \multicolumn{3}{c}{\textbf{1440P}} \\
    & Lat $\downarrow \ $ & TP $\uparrow \ $ & Mem $\downarrow \ $ & Lat $\downarrow \ $ & TP $\uparrow \ $ & Mem $\downarrow \ $ & Lat $\downarrow \ $ & TP $\uparrow \ $ & Mem $\downarrow \ $ & Lat $\downarrow \ $ & TP $\uparrow \ $ & Mem $\downarrow \ $ \\
    \midrule
    Depth Anything V2 \cite{depth_anything_v2}
        & 18.6   & 53.7   & 1.3   & 33.9   & 29.5   & 1.5   & 80.6    & 12.4  & 1.7   & 183.5   & 5.5   & 2.1 \\
    Depth Anything 3 \cite{depthanything3}
        & 14.9   & 67.0   & 1.3   & 58.1   & 17.2   & 1.3   & 141.2   & 7.1   & 1.3   & 207.6   & 4.8   & 1.4 \\
    MoGe-2 \cite{wang2025moge2}
        & 24.1   & 41.5   & 1.3   & 49.3   & 20.3   & 1.5   & 115.3   & 8.7   & 1.7   & 245.5   & 4.1   & 2.1 \\
    Marigold v1.1 \cite{ke2025marigold}
        & 119.1  & 8.4    & 6.0   & 328.1  & 3.0    & 8.0   & 758.1   & 1.3   & 11.7  & 1479.0  & 0.7   & 16.9\\
    Lotus-2 \cite{he2025lotus}
        & 1258.6 & 0.8    & 37.5  & 3095.3 & 0.3    & 39.5  & 6952.7  & 0.1   & 43.3  & 13747.0 & 0.1   & 48.5\\
    Pixel-Perfect Depth \cite{xu2025pixel}
        & 75.3   & 13.3   & 3.2   & 250.2  & 4.0    & 3.3   & 905.5   & 1.1   & 3.5   & 2457.0  & 0.4   & 3.8 \\
    \midrule
    \textbf{Lapis (Ours)}
        & 27.5   & 36.4   & 3.3   & 55.4   & 18.0   & 3.4   & 119.0   & 8.4   & 3.6   & 224.7   & 4.5   & 3.9 \\
    \bottomrule
    \end{tabular}%
  }
\end{table}

To quantify the details of the estimated depth maps, we evaluate the boundary sharpness F1-score proposed by \cite{Bochkovskii2024:arxiv} across three high-resolution synthetic benchmarks: Hypersim \cite{roberts:2021} (indoor), Sintel \cite{Butler:ECCV:2012_Sintel} (various), and Spring \cite{Mehl2023_Spring} (outdoor).
Unlike global error metrics, this evaluation specifically measures the model's ability to resolve sharp occlusion boundaries and fine-grained structural transitions.
As demonstrated in \cref{tab:comp_boundary_sharpness}, Lapis consistently outperforms both discriminative and generative baselines in boundary sharpness across all test resolutions, validating the efficacy of our PRM in recovering high-frequency spatial details.

\subsection{Runtime Analysis}

We evaluate the runtime performance of Lapis against competitive methods on a single NVIDIA RTX Pro 6000 Blackwell GPU with 1 sample per batch.
As summarized in \cref{tab:comp_efficiency}, Lapis significantly outperforms all generative baselines in terms of inference latency.
By leveraging a one-step diffusion formulation and linearized attention, our model achieves a processing speed that is highly competitive with purely regression-based architectures.

\subsection{Ablation Study}

To facilitate efficient analysis, we follow \cite{wang2025moge2, depthanything3} to adopt a lightweight configuration based on the DiT-Base \cite{Peebles2022DiT} backbone.
For the PRM, we reduce the complexity to $M = 2$ blocks with an embedding dimension of $D_{\text{pixel}} = 16$. 

\subsubsection{Effectiveness of components.}

We conduct a component-wise ablation to validate the synergy of the PCM and the PRM, as reported in \cref{tab:ablation_component-wise} and \cref{fig:component-wise_visualization}.
First, removing Global PCM causes a sharp accuracy drop, confirming the necessity of semantic supervision for global coherence.
Second, Local PCM is critical for inter-patch continuity. Its absence increases AbsRel ($4.8 \to 5.1$) and introduces grid artifacts. 
Finally, while the PRM has a marginal impact on accuracy metrics, it is essential for restoring high-frequency details otherwise diluted in the deep linearized backbone to ensure detailed geometry.

\begin{table}[t]
  \caption{
      \textbf{Ablation study of Lapis components.} 
      We evaluate the contribution of each module across five datasets.
  }
  \label{tab:ablation_component-wise}
  \centering
  \resizebox{\linewidth}{!}{%
    \begin{tabular}{l  cc  cc  cc  cc  cc  cc}
    \toprule
    \multirow{2}{*}{\textbf{Components}} & \multicolumn{2}{c}{\textbf{NYUv2}} & \multicolumn{2}{c}{\textbf{KITTI}} & \multicolumn{2}{c}{\textbf{ETH3D}} & \multicolumn{2}{c}{\textbf{ScanNet}} & \multicolumn{2}{c}{\textbf{DIODE}} & \multicolumn{2}{c}{\textbf{Average}} \\
    & AbsRel$\downarrow$ & $\ \ \delta_1\uparrow \ \ $ & AbsRel$\downarrow$ & $\ \ \delta_1\uparrow \ \ $ & AbsRel$\downarrow$ & $\ \ \delta_1\uparrow \ \ $ & AbsRel$\downarrow$ & $\ \ \delta_1\uparrow \ \ $ & AbsRel$\downarrow$ & $\ \ \delta_1\uparrow \ \ $ & AbsRel$\downarrow$ & $\ \ \delta_1\uparrow \ \ $ \\
    \midrule
    Full Model                      & 4.3  & 98.0 & 5.7  & 96.9 & 4.0  & 98.6 & 4.5  & 97.7 & 5.6  & 97.0 & 4.8  & 97.7 \\
    \midrule
    --- \textit{w/o} Global PCM    & 19.6 & 68.2 & 20.2 & 69.9 & 19.1 & 71.4 & 19.1 & 69.1 & 21.2 & 71.9 & 19.8 & 70.1 \\
    --- \textit{w/o} Local PCM     & 4.5  & 97.7 & 6.4  & 96.5 & 4.3  & 98.5 & 4.5  & 97.7 & 5.6  & 97.0 & 5.1  & 97.5 \\
    --- \textit{w/o} PRM           & 4.5  & 97.8 & 5.7  & 96.9 & 4.0  & 98.6 & 4.8  & 97.6 & 5.5  & 97.1 & 4.9  & 97.6 \\
    \bottomrule
    \end{tabular}%
  }
\end{table}
\begin{figure}[tb]
  \centering
  \includegraphics[
    width=\textwidth,
    trim=2cm 5.5cm 2cm 7.2cm,
    clip
  ]{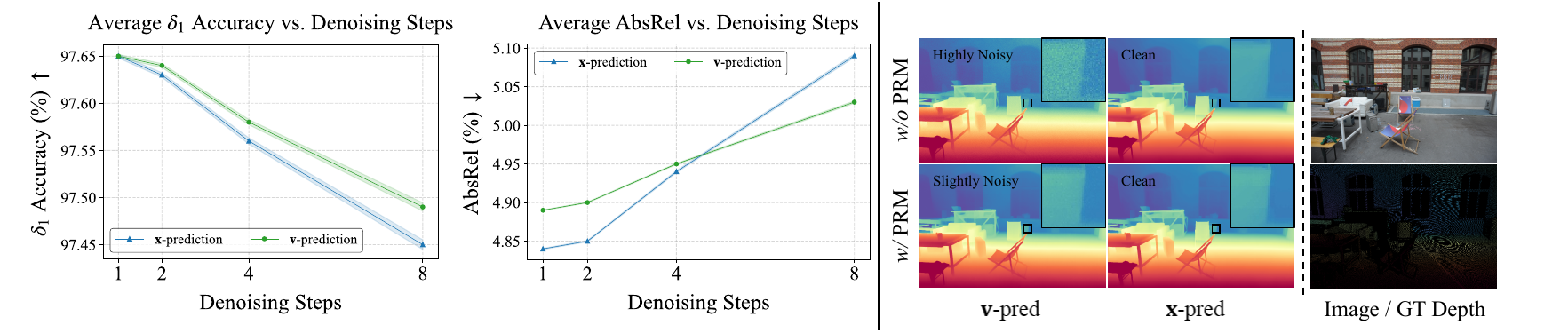}
  \caption{
    \textbf{Quantitative comparison under various denoising steps and qualitative comparison for the impact of the PRM on one-step generation.}
    A one-step $\mathbf{x}$-prediction strategy is optimal in both efficiency and geometric fidelity.
    }
  \label{fig:diffusion_strategy}
\end{figure}

\subsubsection{Prediction targets and diffusion steps.}

We compare $\mathbf{x}$-prediction with $\mathbf{v}$-prediction across varying sampling steps ($1, 2, 4, 8$).
As shown in \cref{fig:diffusion_strategy} (left), both strategies peak at one-step sampling, with $\mathbf{x}$-prediction slightly outperforming in AbsRel. 
We further analyze pixel-level continuity. 
As visualized in \cref{fig:diffusion_strategy} (right), while the PRM effectively suppresses noise in the $\mathbf{v}$-prediction path, $\mathbf{x}$-prediction consistently yields more spatially coherent depth maps. 
Consequently, a one-step $\mathbf{x}$-prediction strategy is identified as our optimal choice, as it inherently aligns with the clean data manifold and maximizes both efficiency and geometric fidelity.

\section{Conclusion}

In this paper, we presented Lapis, a pixel-space generative model based on linear attention for efficient and high-quality depth estimation, especially under high-resolution inputs.
By decoupling patch-level context modeling from pixel-level refinement and leveraging a direct $\mathbf{x}$-prediction strategy, we effectively resolve the degradations in the naively linearized backbone and suppress the noise in one-step pixel-space diffusion.
Extensive experiments demonstrate that Lapis achieves SOTA zero-shot performance while significantly reducing inference latency.
Our approach provides a scalable and computationally efficient foundation for real-world 3D perception, bridging the gap between generative quality and discriminative speed.

\section*{Acknowledgements}

This work was supported by the National Natural Science Foundation of China (Grant Nos. 62322113, 62376156), as well as the Shanghai Municipal Special Program for Basic Research on General AI Foundation Models (Grant No. 2025SHZDZX025G15).

\bibliographystyle{splncs04}
\bibliography{main}

\clearpage
\appendix
\renewcommand{\thesection}{\Alph{section}}
\renewcommand{\thesubsection}{\thesection.\arabic{subsection}}
\renewcommand{\thefigure}{\Alph{figure}}
\renewcommand{\thetable}{\Alph{table}}

\section{More Implementation Details}

\Cref{tab:model_specs} shows the hyperparameters of Lapis and its ablation variant.
Both configurations are designed to remain strictly comparable with DiT-L and DiT-B \cite{Peebles2022DiT} in hidden dimensions and the number of blocks, respectively.

\begin{table}[th]
  \vspace{-10pt}
  \caption{
      \textbf{Detailed architectural configurations for Lapis.} 
  }
  \label{tab:model_specs}
  \centering
  \vspace{-5pt}
  \setlength{\tabcolsep}{5pt}
  \resizebox{\linewidth}{!}{%
    \begin{tabular}{l @{\extracolsep{\fill}} ccccc c ccc}
    \toprule
    \multirow{2}{*}{\textbf{Model}} & \multicolumn{5}{c}{\textbf{Linearized DiT Backbone}} & \multicolumn{1}{c}{\textbf{PCM}} & \multicolumn{3}{c}{\textbf{PRM}} \\
    \cmidrule(lr){2-6} \cmidrule(lr){7-7} \cmidrule(lr){8-10}
    & Patch Size & Blocks & Embed Dim & Attn Heads & FFN Dim & DINOv2 & Blocks & Embed Dim & FFN Dim \\
    \midrule
    Lapis (Full)     & $16 \times 16$ & 24 & 1024 & 16 & 4096 & ViT-L/14 & 3 & 32 & 128 \\
    Lapis (Ablation) & $16 \times 16$ & 12 & 768  & 12 & 3072 & ViT-B/14 & 2 & 16 & 64  \\
    \bottomrule
    \end{tabular}%
  }
  \vspace{-25pt}
\end{table}

\section{Evaluation Protocols}

\subsection{Relative Depth Estimation}

For zero-shot relative depth evaluation, we adopt the open-source protocol from Marigold \cite{ke2023repurposing, ke2025marigold} for pre-processing, least-squares alignment, and metric computation.
Additionally, we downsample ETH3D \cite{schoeps2017cvpr} to $2048 \times 1360$, and use a Sobel filter with a gradient threshold of $0.3$ to mask out unreliable depth transitions caused by sensor noise in DIODE \cite{diode_dataset}.
All models perform inference at their respective scales (\eg, native, 1080P, or 1440P), with predictions resized to ground-truth resolution before alignment and metric computation.

\subsection{Boundary Sharpness}

For boundary sharpness evaluation, we adopt the open-source protocol from Depth Pro \cite{Bochkovskii2024:arxiv} for metric computation.
Consistent with our relative depth evaluation, all models perform inference at their respective scales, with the results subsequently resized to the ground-truth resolution.
These predictions are then aligned via least-squares and clipped to the valid depth range of each dataset before computing the final sharpness metrics.

\subsection{Runtime Analysis}

We evaluate the runtime efficiency on a single NVIDIA RTX Pro 6000 Blackwell GPU with a batch size of 1.
We incorporate Memory Efficient Attention \cite{xFormers2022} for standard Softmax attention \cite{NIPS2017_3f5ee243} operations, and further optimize all models via compilation to minimize execution overhead.
The reported results are the average of $100$ iterations after $20$ warm-up rounds to eliminate cold-start overhead.
Notably, Lapis significantly outperforms existing generative models in efficiency even without specialized CUDA kernel optimizations for linear attention.

\section{Edge-Aware Point Cloud Evaluation}

\begin{table}[t]
  \caption{
      \textbf{Edge-aware point cloud evaluation on the test split of Hypersim \cite{roberts:2021}.} 
      We report the log-space Chamfer Distance (CD) near depth discontinuities.
  }
  \label{tab:chamfer_distance}
  \centering
  \resizebox{\linewidth}{!}{%
      \setlength{\tabcolsep}{3pt}
      \begin{tabular}{c cccccccccc}
        \toprule
        \textbf{Method} & Depth Any. V2 \cite{depth_anything_v2} & Depth Any. 3 \cite{depthanything3} & MoGe-2 \cite{wang2025moge2} & Marigold v1.1 \cite{ke2025marigold} & PPD \cite{xu2025pixel} & \textbf{Lapis (Ours)} \\
        \midrule
        $\textbf{Log-Space CD} \downarrow$ & 0.19 & 0.17 & 0.14 & 0.18 & 0.12 & \textbf{0.12} \\
        \bottomrule
      \end{tabular}
  }
  \vspace{-10pt}
\end{table}

\begin{figure}[tb]
  \centering
  \includegraphics[
    width=\textwidth,
    trim=0cm 0cm 3.25cm 6cm,
    clip
  ]{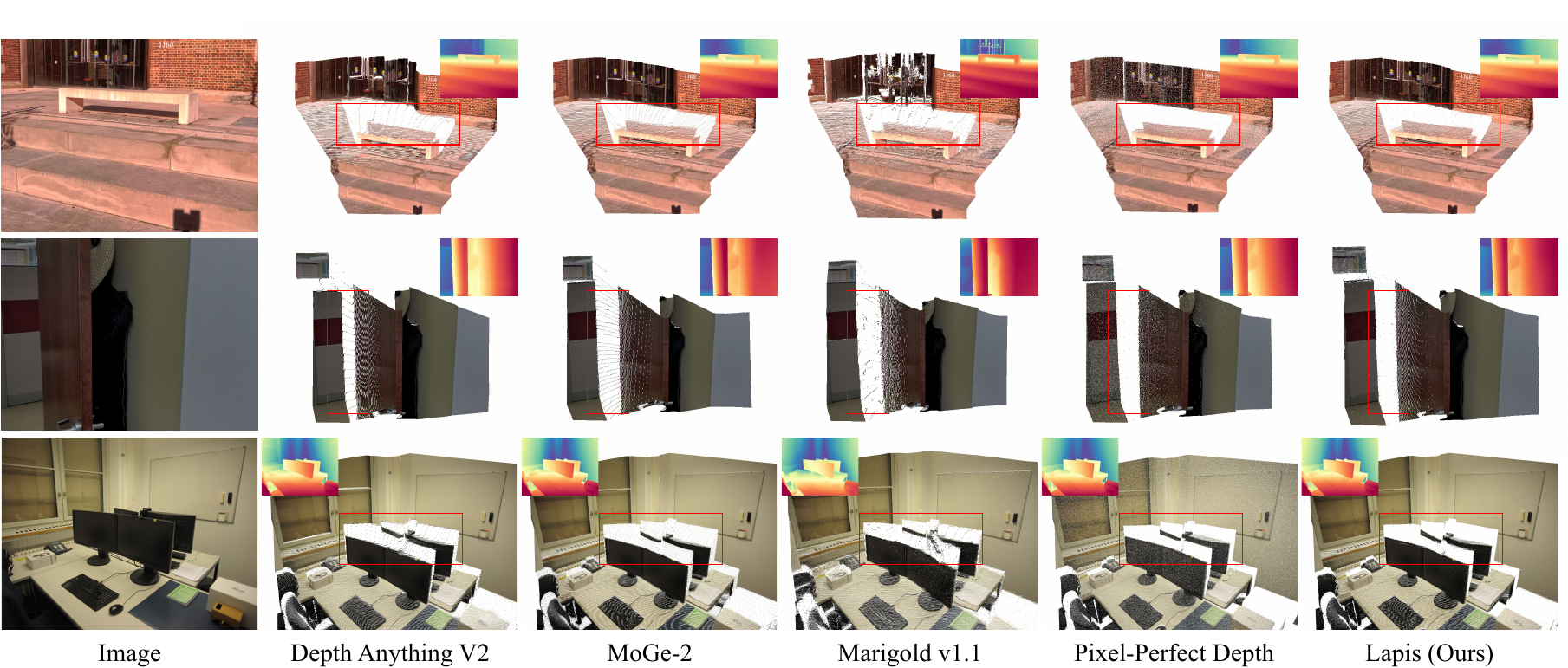}
  \vspace{-15pt}
  \caption{
    \textbf{Qualitative comparisons for point clouds without post-processing.}
    Compared to discriminative \cite{depth_anything_v2, wang2025moge2} and latent-space generative \cite{ke2025marigold} methods, Lapis significantly suppresses flying pixel artifacts (highlighted in red boxes).
    Furthermore, while previous pixel-space generative models \cite{xu2025pixel} often suffer from ``snow-like'' noise in flat regions, Lapis produces more spatially consistent surfaces.
    \textit{Best viewed zoomed-in.}
    }
  \vspace{-10pt}
  \label{fig:pointcloud}
\end{figure}

While current discriminative and latent-space generative models often suffer from flying pixel artifacts due to over-smoothing or the lossy compression of VAEs \cite{kingma2022autoencodingvariationalbayes}, Lapis preserves uncompromising geometric fidelity as a pixel-space diffusion framework.
To validate this, we follow \cite{xu2025pixel} to evaluate edge-aware point clouds on the official test split of Hypersim \cite{roberts:2021}.
Unlike the boundary F1-score, this evaluation provides a more rigorous assessment from the perspective of 3D structural integrity.
Specifically, predicted depth maps are first aligned with the ground truth and back-projected into 3D point clouds with camera intrinsics.
We then apply a Canny detector to the ground-truth depth maps in log-space to extract edge masks and subsequently compute the log-space Chamfer Distance (CD) near these regions.
As demonstrated in \cref{tab:chamfer_distance}, Lapis achieves the lowest log-space CD (0.12), matching the SOTA pixel-space generative model \cite{xu2025pixel} and significantly outperforming discriminative and latent-space generative ones \cite{depth_anything_v2, depthanything3, wang2025moge2, ke2025marigold}.
The visualization in \cref{fig:pointcloud} further highlights that Lapis produces smoother and more spatially consistent surfaces.

\section{Extended Ablation Study}

\begin{table}[t]
  \caption{
      \textbf{Ablation study of different kernel functions.} 
      Our linearized configurations maintain performance parity with the standard Softmax attention, while the $\text{ELU} + 1$ kernel yields a slightly superior overall result.
  }
  \label{tab:ablation_kernel_function}
  \centering
  \vspace{-5pt}
  \resizebox{\linewidth}{!}{%
    \begin{tabular}{l  cc  cc  cc  cc  cc  cc}
    \toprule
    \multirow{2}{*}{\textbf{Kernel} $\quad$} & \multicolumn{2}{c}{\textbf{NYUv2}} & \multicolumn{2}{c}{\textbf{KITTI}} & \multicolumn{2}{c}{\textbf{ETH3D}} & \multicolumn{2}{c}{\textbf{ScanNet}} & \multicolumn{2}{c}{\textbf{DIODE}} & \multicolumn{2}{c}{\textbf{Average}} \\
    & AbsRel$\downarrow$ & $\ \ \delta_1\uparrow \ \ $ & AbsRel$\downarrow$ & $\ \ \delta_1\uparrow \ \ $ & AbsRel$\downarrow$ & $\ \ \delta_1\uparrow \ \ $ & AbsRel$\downarrow$ & $\ \ \delta_1\uparrow \ \ $ & AbsRel$\downarrow$ & $\ \ \delta_1\uparrow \ \ $ & AbsRel$\downarrow$ & $\ \ \delta_1\uparrow \ \ $ \\
    \midrule
    Softmax (Baseline)      & 4.6  & 98.0 & 5.7  & 96.9 & 4.1  & 98.6 & 5.0  & 97.6 & 5.6  & 97.1 & 5.0  & 97.6 \\
    \midrule
    $\text{ReLU}$           & 4.5  & 97.9 & 5.7  & 97.0 & 4.0  & 98.7 & 4.9  & 97.6 & 5.5  & 97.1 & 4.9  & 97.7 \\
    $\text{Focused ReLU}, p=3$   & 4.5  & 97.9 & 5.7  & 96.9 & 3.9  & 98.7 & 4.7  & 97.7 & 5.6  & 97.0 & 4.9  & 97.6 \\
    $\text{ELU} + 1$        & 4.3  & 98.0 & 5.7  & 96.9 & 4.0  & 98.6 & 4.5  & 97.7 & 5.6  & 97.0 & 4.8  & 97.7 \\
    \bottomrule
    \end{tabular}%
  }
  \vspace{-5pt}
\end{table}

\begin{table}[t]
  \caption{
      \textbf{Ablation study of semantic conditioning strategies.} 
      Methods enforcing a direct one-to-one spatial correspondence (MLP and AdaLN variants) significantly outperform global Linear Cross-Attention in preserving the geometric layout.
  }
  \label{tab:ablation_conditioning}
  \centering
  \vspace{-5pt}
  \resizebox{\linewidth}{!}{%
    \begin{tabular}{l  cc  cc  cc  cc  cc  cc}
    \toprule
    \multirow{2}{*}{\textbf{Conditioning} $\quad$} & \multicolumn{2}{c}{\textbf{NYUv2}} & \multicolumn{2}{c}{\textbf{KITTI}} & \multicolumn{2}{c}{\textbf{ETH3D}} & \multicolumn{2}{c}{\textbf{ScanNet}} & \multicolumn{2}{c}{\textbf{DIODE}} & \multicolumn{2}{c}{\textbf{Average}} \\
    & AbsRel$\downarrow$ & $\ \ \delta_1\uparrow \ \ $ & AbsRel$\downarrow$ & $\ \ \delta_1\uparrow \ \ $ & AbsRel$\downarrow$ & $\ \ \delta_1\uparrow \ \ $ & AbsRel$\downarrow$ & $\ \ \delta_1\uparrow \ \ $ & AbsRel$\downarrow$ & $\ \ \delta_1\uparrow \ \ $ & AbsRel$\downarrow$ & $\ \ \delta_1\uparrow \ \ $ \\
    \midrule
    Linear Cross-Attention  & 10.3 & 89.6 & 13.9 & 82.7 & 9.8  & 91.2 & 9.7  & 90.5 & 10.4 & 91.4 & 10.8 & 89.1 \\
    \midrule
    MLP                     & 4.4  & 98.0 & 5.8  & 96.9 & 4.0  & 98.7 & 4.6  & 97.8 & 5.6  & 97.0 & 4.9  & 97.7 \\
    AdaLN (Int.)            & 4.5  & 97.9 & 5.8  & 96.9 & 4.0  & 98.7 & 4.7  & 97.7 & 5.6  & 97.1 & 4.9  & 97.7 \\
    AdaLN-Zero (Full)       & 4.4  & 97.9 & 5.8  & 96.7 & 4.0  & 98.6 & 4.8  & 97.5 & 5.6  & 97.1 & 4.9  & 97.6 \\
    AdaLN-Zero (Int.)       & 4.3  & 98.0 & 5.7  & 96.9 & 4.0  & 98.6 & 4.5  & 97.7 & 5.6  & 97.0 & 4.8  & 97.7 \\
    \bottomrule
    \end{tabular}%
  }
  \vspace{-5pt}
\end{table}

In this section, we additionally investigate the architectural choices of Lapis, focusing on linear attention kernels and semantic conditioning strategies.

\subsection{Influence of Kernel Functions}

We evaluate various activations within the linearized DiT backbone of the complete Lapis architecture, with a standard Softmax attention variant as a baseline for comparison.
Specifically, we contrast the ReLU kernel adopted in \cite{xie2024sana}, the $\text{ELU} + 1$ kernel as the canonical setting for linear attention \cite{katharopoulos2020transformers}, and the Focused ReLU kernel \cite{han2023flatten}, which enhances the focus ability of ReLU via a power function $f_p(x) = \frac{\|x\|}{\|x^p\|}x^p$.
As shown in Table \ref{tab:ablation_kernel_function}, the linearized models achieve performance competitive with, or even slightly exceeding, the Softmax baseline, validating the effectiveness of our architectural design.
Among them, ELU + 1 yields slightly superior average performance and is thus selected as our final configuration.

\subsection{Semantic Conditioning Strategies}

We investigate several mechanisms for semantic conditioning inside the PCM, drawing inspiration from the class-label conditioning used in DiT \cite{Peebles2022DiT}.
These include: 
1) Linear Cross-Attention, which treats semantic features as keys and values for global interaction; 
2) a simple MLP that concatenates features with image tokens; 
and 3) AdaLN variants that modulate features through adaptive layer normalization.
As shown in \cref{tab:ablation_conditioning}, Linear Cross-Attention yields significantly inferior results.
We attribute this to the absence of a patch-wise inductive bias.
Specifically, cross-attention allows each token to aggregate information from the entire semantic map, which blurs the precise spatial alignment between semantic tokens and DiT tokens, and compromises the global geometric layout that the PCM is intended to regularize.
In contrast, all methods that preserve this localized inductive bias achieve consistently superior and comparable performance.
Among them, interleaved AdaLN-Zero is selected as our final configuration, as it offers slightly better average results without introducing additional parameters.

\section{More Visual Results}

\begin{figure}[tb]
  \centering
  \includegraphics[
    width=\textwidth,
    trim=0cm 0cm 0cm 6.5cm,
    clip
  ]{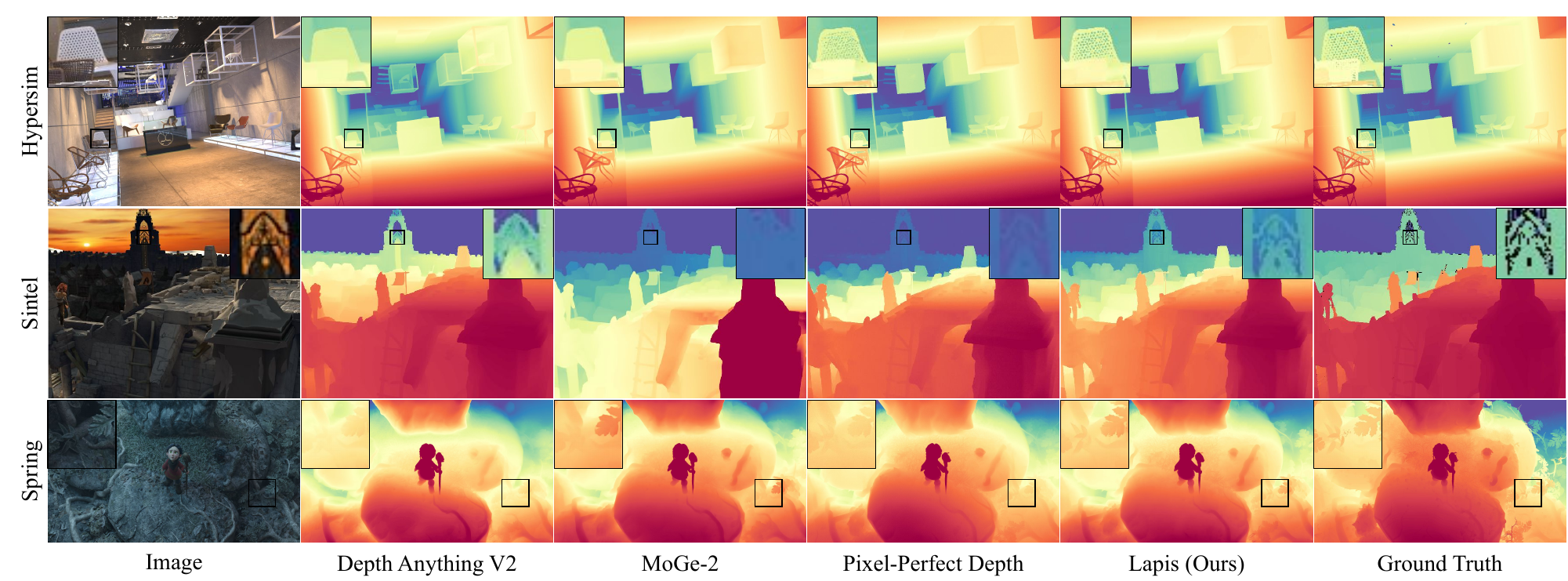}
  \vspace{-15pt}
  \caption{
    \textbf{Qualitative comparisons on three synthetic benchmarks.}
    We visualize depth estimation results in Hypersim \cite{roberts:2021}, Sintel \cite{Butler:ECCV:2012_Sintel}, and Spring \cite{Mehl2023_Spring}.
    Lapis demonstrates robust performance in preserving sharp structural boundaries across diverse scenarios, ranging from complex indoor geometries to wide-scale outdoor landscapes.
    }
  \label{fig:synthetic_visualization}
\end{figure}

\begin{figure}[t]
  \centering
  \includegraphics[
    width=\textwidth,
    trim=0cm 0cm 0cm 2cm,
    clip
  ]{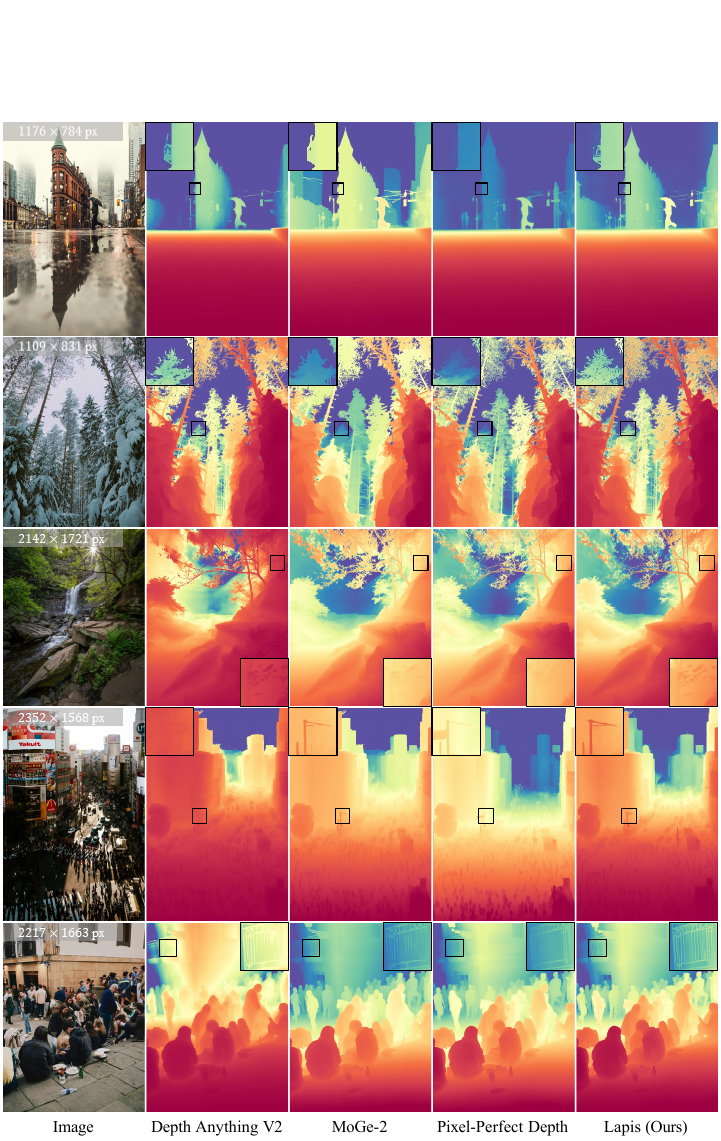}
  \vspace{-15pt}
  \caption{
    \textbf{Qualitative results on open-domain images.}
    \textit{Best viewed zoomed-in.}
    }
  \vspace{-10pt}
  \label{fig:open-domain_visualization}
\end{figure}

We provide additional qualitative comparisons on synthetic benchmarks, including Hypersim \cite{roberts:2021}, Sintel \cite{Butler:ECCV:2012_Sintel}, and Spring \cite{Mehl2023_Spring}, as well as open-domain images sourced from publicly available web resources, \eg, Pexels \footnote{\url{https://www.pexels.com/}} and Unsplash \footnote{\url{https://unsplash.com/}}, in \cref{fig:synthetic_visualization} and \cref{fig:open-domain_visualization}, respectively.
These results further demonstrate the robustness and structural fidelity of Lapis across diverse scenarios.

\section{Limitations and Future Work}

Currently, linear attention exhibits numerical instability in reduced-precision formats (\eg, fp16).
This challenge arises from the lack of explicit normalization in the attention score computation process.
Addressing this through more stable formulations and optimized kernels will enhance both numerical robustness and throughput.
Furthermore, Lapis presently focuses on single-image relative depth estimation.
Extending this framework to metric depth or video-consistent estimation would significantly broaden its utility in downstream applications, such as robotics and temporal-coherent scene reconstruction.

\clearpage  


\end{document}